\documentclass[10pt,journal,compsoc]{IEEEtran}

\usepackage{diagbox}
\usepackage{subcaption}
\usepackage{comment}
\usepackage{amsmath,amssymb} 
\usepackage{color}
\usepackage{dsfont}
\usepackage{booktabs}
\usepackage{multirow}
\usepackage{xcolor}
\usepackage{colortbl}
\usepackage{graphicx}
\usepackage{paralist}
\usepackage{marvosym}
\usepackage[export]{adjustbox}
\usepackage{makecell}

\newcommand\sotaa{\textcolor{red}}
\newcommand\sotab{\textcolor{blue}}

\definecolor{sh_gray}{rgb}{0.84,0.84,0.84}
\definecolor{sh_gray2}{rgb}{1,0.89,0.75}
\definecolor{color3}{rgb}{0.95,0.95,0.95}
\definecolor{color5}{rgb}{0.90,0.90,0.90}
\definecolor{color4}{rgb}{0.94,0.94,1}

\newcolumntype{M}[1]{>{\centering\arraybackslash}m{#1}}

\makeatother
\usepackage[pagebackref=true,breaklinks=true,letterpaper=true,colorlinks,bookmarks=false,citecolor=blue,linkcolor=blue,urlcolor=blue]{hyperref}

\ifCLASSOPTIONcompsoc
  \usepackage[nocompress]{cite}
\else
  \usepackage{cite}
\fi

\begin{document}
\title{
DeepRFTv2: Kernel-level Learning for Image Deblurring
}

\iftrue
\author{
		Xintian~Mao,
        Haofei~Song,
        Yin-Nian Liu,~\IEEEmembership{Member,~IEEE},\\
        Qingli~Li,~\IEEEmembership{Senior Member,~IEEE},
        Yan~Wang,~\IEEEmembership{Member,~IEEE}
\thanks{
This work was supported by the National Natural Science Foundation of China (Grant No.  62471182, 62475072), Fundamental Research Funds for the Central Universities, Shanghai Rising-Star Program (Grant No. 24QA2702100), the Science and Technology Commission of Shanghai Municipality (Grant No. 22S31905800, 25JD1401300, 22DZ2229004)
}
\IEEEcompsocitemizethanks{
\IEEEcompsocthanksitem Xintian Mao, Haofei Song, Qingli Li and Yan Wang are with Shanghai Key Laboratory of Multidimensional Information Processing, East China Normal University, Shanghai, China.
\IEEEcompsocthanksitem Yin-Nian Liu is with the State Key Laboratory of Infrared Physics, Shanghai Institute of Technical Physics, Chinese Academy of Sciences, Shanghai, China, and also with the University of Chinese Academy of Sciences, Beijing, China.
}
\thanks{
$\textrm{\Letter}$ Corresponding author:
Yan~Wang (ywang@cee.ecnu.edu.cn)
}
}
\fi

%
%

\markboth{
}%
{Shell \MakeLowercase{\textit{et al.}}: Bare Demo of IEEEtran.cls for Computer Society Journals}
\IEEEtitleabstractindextext{%
\begin{abstract}
It is well-known that if a network aims to learn how to deblur, it should understand the blur process. Blurring is naturally caused by the convolution of the sharp image with the blur kernel. Thus, allowing the network to learn the blur process in the kernel-level can significantly improve the image deblurring performance. But, current deep networks are still at the pixel-level learning stage, either performing end-to-end pixel-level restoration or stage-wise pseudo kernel-level restoration, failing to enable the deblur model to understand the essence of the blur. 
To this end, we propose Fourier Kernel Estimator (FKE), which considers the activation operation in Fourier space and converts the convolution problem in the spatial domain to a multiplication problem in Fourier space. Our FKE, jointly optimized with the deblur model, enables the network to learn the kernel-level blur process with low complexity and without any additional supervision. Furthermore, we change the convolution object of the kernel from ``image" to network extracted ``feature", whose rich semantic and structural information is more suitable to blur process learning. With the convolution of the feature and the estimated kernel, our model can learn the essence of blur in kernel-level. 
To further improve the efficiency of feature extraction, we design a decoupled multi-scale architecture with multiple hierarchical sub-unets with a reversible strategy, which allows better multi-scale encoding and decoding in low training memory. Extensive experiments indicate that our method achieves state-of-the-art motion deblurring results and show potential for handling other kernel-related problems. Analysis also shows our kernel estimator is able to learn physically meaningful kernels. The code will be available at \href{https://github.com/DeepMed-Lab-ECNU/Single-Image-Deblur}{https://github.com/DeepMed-Lab-ECNU/Single-Image-Deblur}.

\end{abstract}

}

\maketitle

\IEEEdisplaynontitleabstractindextext

%
\IEEEpeerreviewmaketitle


%
%
%
%

\begin{figure*}[ht]
\begin{center}
    \includegraphics[width=0.9\linewidth]{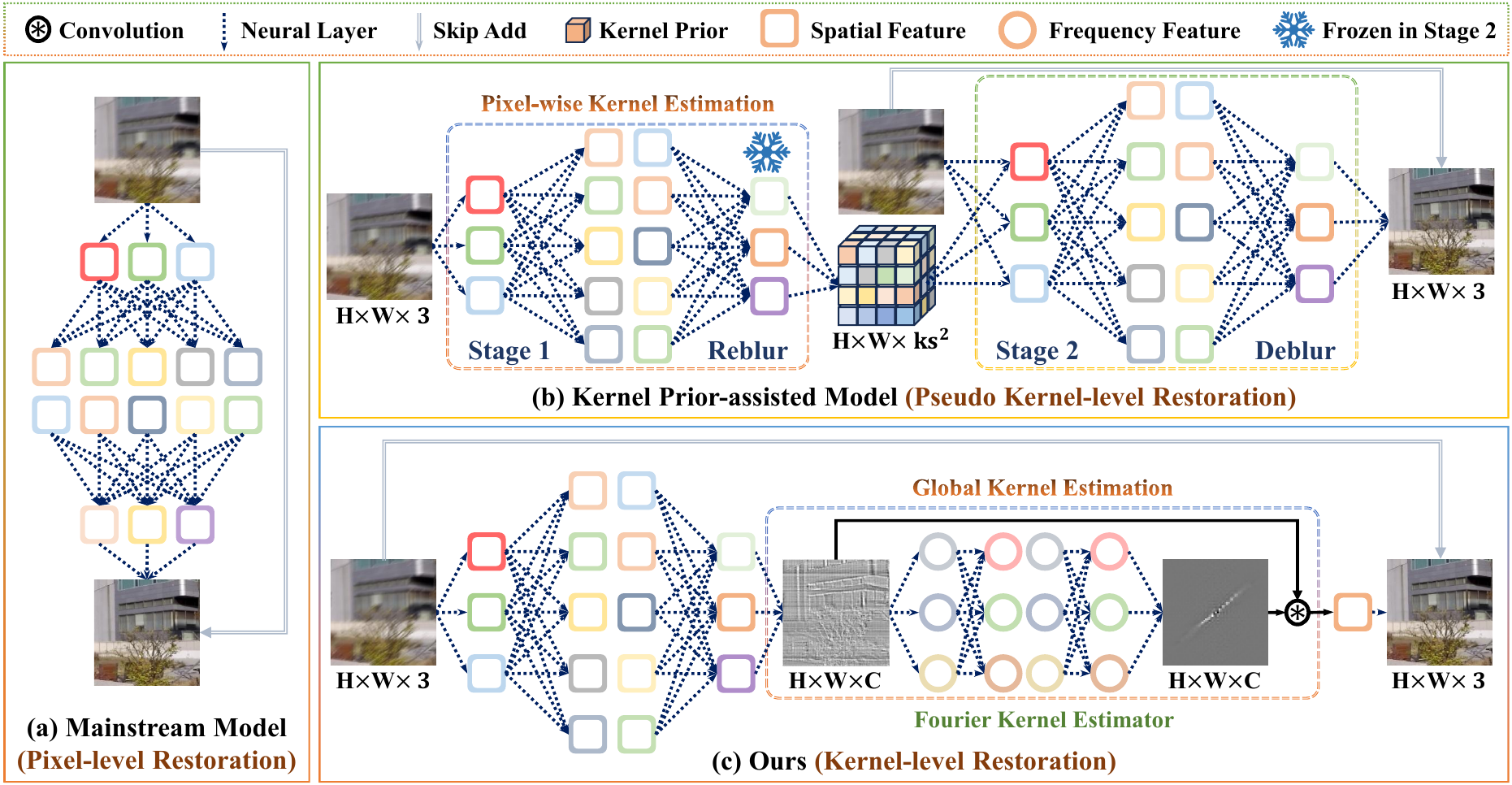}
\end{center}
\vspace{-1.em}
\caption{Three deblurring model architectures.
(a) One-stage deblurring models~\cite{Nah2017deep, Cho2021rethinking, Zamir2021restormer, Chen2022simple} directly predict blur patterns using neural networks in the spatial domain under an end-to-end training scheme.
(b) Kernel prior-assisted models~\cite{fang2023UFPNet, Fang_PGDN} employ a two-stage strategy: first pre-training a reblur model to obtain kernel priors, then incorporating these priors as spatial features into the deblurring network.
(c) Our method integrates the deblurring model and kernel estimator into a unified end-to-end framework, where the estimated kernels are physically applied in the convolution process.}
\label{fig:DeepRFTv2}
\vspace{-1.em}
\end{figure*}
\IEEEraisesectionheading{\section{Introduction}\label{sec:introduction}}
\IEEEPARstart{I}{mage} deblurring, as a subtask of image restoration, aims at restoring the sharp image $\mathrm{S}$ from the blur image $\mathrm{B}$. Generally, the degradation of $\mathrm{B}$ is caused by the convolution of $\mathrm{S}$ with the blur kernel $\mathrm{k}$, and additive noise $\mathrm{n}$:
\begin{equation}
\mathrm{B} = \mathrm{S} * \mathrm{k} + \mathrm{n},
\label{eq:blur}
\end{equation}
where $*$ indicates the convolution operator. 
The blur kernel $\mathrm{k}$ is unknown in practice. Thus, any blind deblurring method, by nature, requires explicit or implicit estimation of the blur kernel $\mathrm{k}$ to restore the sharp image.

To restore the sharp image from the blur image, traditional blind deblurring methods~\cite{fergus2006removing,joshi2008psf,cho2011handling,pan2016blind,yan2017image,whyte2012non,xu2013unnatural} tend to estimate the blur kernel and the latent sharp image within the Maximum A Posteriori (MAP) framework~\cite{Pan2019phase}. 
Later, the pioneering deep learning-based~\cite{Bahat_2017_ICCV, chakrabarti2016neural, Schuler2016learning, Sun2015learning} approaches leveraged convolutional neural networks (CNNs) to estimate the blur kernel and then restored the latent sharp image by the estimated kernel. However, these approaches face several critical challenges: (1) Simplistic CNN architectures are fundamentally limited in handling the complex scene and blur diversity in image deblurring tasks. (2) The estimated kernels are sensitive to noise and other factors, resulting in artifacts in the restored image. (3) Estimating the spatially variant kernel for each pixel in dynamic scenes entails unbearable computational and memory costs~\cite{Nah2017deep}.

To avoid the above problems, almost all recent end-to-end deep learning-based methods~\cite{Nah2017deep,Tao2018scale, Kupyn2018deblurgan,Kupyn2019deblurgan, Lin2019TellMW,Wang2022uformer,Zamir2021restormer} (Fig.~\ref{fig:DeepRFTv2} (a)) tend to use the network to learn a direct pixel-wise mapping: $\mathcal{F}: \mathrm{B} \rightarrow \mathrm{P}$, which predicts the blur pattern $\mathrm{P}$ (the residual of the blur image and its sharp counterpart) from the blur image $\mathrm{B}$ in recent years. 
However, these methods overlook the convolutional nature of the blur degradation (Eq.~\ref{eq:blur}).

\textbf{It is well-known that if a network aims to learn how to deblur, it should understand the blur process.} The recent end-to-end models can be regarded as \textbf{pixel-level restoration}. Since image blur typically arises from the kernel, allowing the network to learn the blur process in the kernel-level can significantly improve the performance.
Compared to the degradation representation learning method~\cite{li2022MSDINet}, several recent kernel prior-assisted methods~\cite{fang2023UFPNet,Fang_PGDN} try to introduce kernel information into the deblur model, which guides the deblur model to understand how the blur occurs. We refer to these methods as \textbf{pseudo kernel-level restoration}. 

As shown in Fig.~\ref{fig:DeepRFTv2} (b), these methods typically contain two stages: first learning blur process for kernel prior, then followed by a deblurring network. 
Physically grounded explicit kernels, supervised by a reblurring process (reblur the sharp image by the estimated kernels), can be learned in these methods in the first stage. 
But in the second stage, these methods often resort to simple fusion strategies (\emph{e.g.}, concatenation or multiplication). They treat the kernel merely as a spatial feature to be combined with blur image features for predicting the blur pattern on a spatial per-pixel basis. This approach fails to simulate the physical blur process (\emph{e.g.}, convolution properties) in the deblur model. 
This fusion choice may stem from the fact: (1) A deconvolution problem is generally not tractable for deep networks; (2) The blur kernel is supervised through an image reblurring process in the first stage during training, where accurately estimating the image-level motion kernels is challenging due to nuisance factors (\emph{e.g.}, noise). Incorrect kernel estimation introduces artifacts into the restored image. Therefore, the kernel prior is only employed as a \textbf{spatial feature} instead of a \textbf{real convolution kernel} for deblurring. Thus, these methods can only be considered as \textbf{pseudo kernel-level restoration}. 


Other limitations of kernel prior-assisted methods include:
(1) The kernel estimation module is usually trained beforehand to avoid computational or optimization challenges if trained jointly with the deblurring network. Thus, the blur process (reblurring) are not learned with the deblur model. (2) The estimated per-pixel kernels, derived from normalizing flow~\cite{fang2023UFPNet} or kernel parameters (length, angle and curvature)~\cite{Fang_PGDN}, are constrained to small sizes. Thus, these methods always fail to estimate the kernel of the same size as the actual kernel.

\begin{figure*}[t]
\begin{center}
    \includegraphics[width=0.88\linewidth]{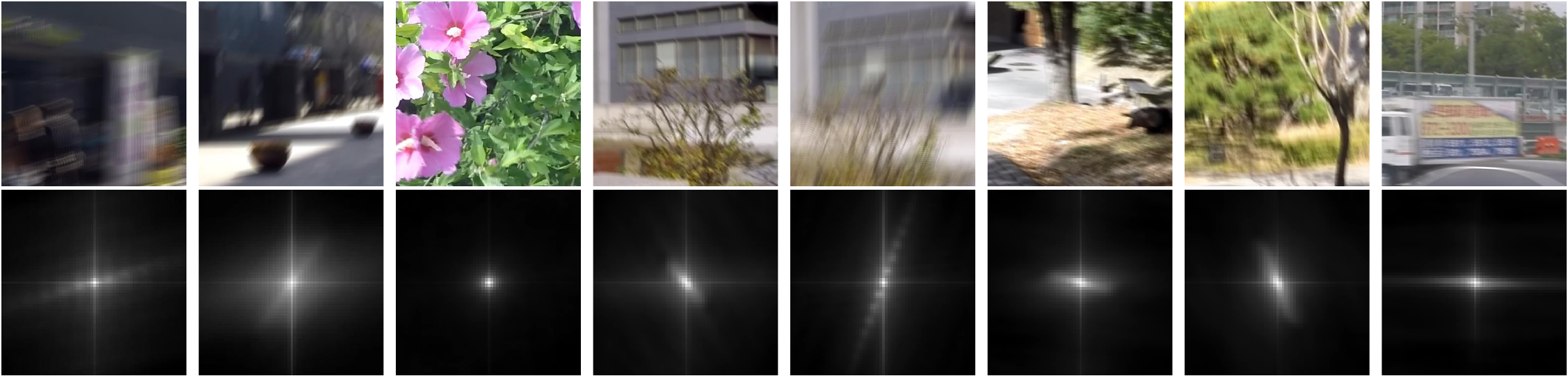}
\end{center}
\vspace{-1.em}
\caption{Visualizations of various blur images $\mathrm{B}$ and ${\sigma}(\mathcal{F}^{-1}(\mathbf{ReLU}(\mathcal{F}(\mathrm{B}))) - \mathrm{B} / 2)$. $\sigma$ denotes a cyclic shift [H/2, W/2].}
\label{fig:fft_relu}
\vspace{-1.3em}
\end{figure*}

To sum up, existing end-to-end deep learning-based deblurring models \cite{Nah2017deep,Tao2018scale} deblur an image via learning pixel-wise blur process (Fig.~\ref{fig:DeepRFTv2} (a)). While kernel prior-assisted methods \cite{fang2023UFPNet,Fang_PGDN} estimate kernels, but do not instantiate physical blur process in the deblur model (Fig.~\ref{fig:DeepRFTv2} (b)). In this paper, we propose an end-to-end deblur model which explores kernel-level blur process learning, so that the deblur model is able to understand the essence of the blur. To this end, we propose to estimate the kernel and directly learn convolution process defined in Eq.~\ref{eq:blur}. We instantiate this process in Fourier space, since: (1) Applying FFT-ReLU-IFFT acts as learning blur mode (\emph{i.e.}, kernel pattern) from a blur image, indicating the blur direction and blur level, as extensively discussed in our previous version DeepRFT \cite{XintianMao2023DeepRFT}. Fusing such kernel information, as shown in Fig.~\ref{fig:fft_relu}, into the network can incorporate kernel priors. (2) Convolution in the spatial domain can be converted into a simpler multiplication in Fourier space, allowing Eq.~\ref{eq:blur} to be rewritten as:
\begin{equation}
    \mathrm{B} = \mathcal{F}^{-1}(\mathcal{F}(\mathrm{S})\odot\mathcal{F}(\mathrm{k})) + \mathrm{n},
\label{eq:fblur}
\end{equation}
where $\mathcal{F}$ and $\mathcal{F}^{-1}$ denote 2D Fast Fourier Transform (FFT2D) and its inverse (IFFT2D)~\cite{cooley1965fft}, while $\odot$ represents Hadamard product. (3) Some blur kernels can be very large in size, making them difficult for networks to learn in the spatial domain. As shown in Fig.~\ref{fig:DeepRFTv2}(c), we refractor the traditional convolution tail in the mainstream model to Fourier Kernel Estimator (FKE), which first transforms the learned feature from the spatial domain to Fourier space, then transforms the feature back to the spatial domain after learning the ``kernel" using a light-weight neural network and conducting the multiplication between the feature and the kernel. Without any supervision, we can learn meaningful ``kernels'' in our FKE module, thanks to the ReLU operation \cite{XintianMao2023DeepRFT} and the multiplication in Fourier space. Our FKE enjoys the benefits that (1) no image-level kernels need to be calculated beforehand, (2) generating global kernels of the same size as the feature, and (3) the estimated kernels are directly convolved
on the features, which allows the network to learn degradation process under a free condition.

Regarding network architecture, a coarse-to-fine network design principle has shown to be effective in image deblurring \cite{Nah2017deep,Tao2018scale}. Conventional coarse-to-fine models employ cascaded sub-networks to progressively restore images, as shown in Fig.~\ref{fig:DMSUNet}(a), but suffer from low restoration efficiency. One-Scale UNet methods \cite{Zamir2021restormer,Chen2022simple} extract multi-scale features within a unified encoder-decoder structure, as shown in Fig.~\ref{fig:DMSUNet}(b), yet they do not explicitly model coarse-to-fine image-level representations. MIMO-UNet \cite{Cho2021rethinking} addresses the above problems by integrating multi-scale inputs and outputs into a single network, where the encoder is modified to accept multi-scale input images ($\mathrm{B}_1$ and $\mathrm{B}_2$ in Fig.~\ref{fig:DMSUNet}(c)) and the decoder is adapted to produce multi-scale sharp outputs ($\mathrm{\hat{S}}_1$ and $\mathrm{\hat{S}}_2$ in Fig.~\ref{fig:DMSUNet}(c)). This design leads to an information aliasing problem: the degradation representation learning can be interfered by accepting low-level features in mid-level encoders (\emph{e.g.}, the encoder which takes $\mathrm{B}_2$ as input in Fig.~\ref{fig:DMSUNet}(c)), while the blur patterns can be decoded prematurely in mid-level decoders. To this end, we refactor MIMO-UNet (Fig.~\ref{fig:DMSUNet} (c)) to a Decoupled Multi-Scale UNet (DMS-UNet) with multiple hierarchical sub-unets, as shown in Fig.~\ref{fig:DMSUNet} (d). Due to the decoupled architecture, each sub-unet can maintain a certain degree of independence during the feature encoding
and decoding process for image deblurring. We name our model DeepRFTv2 due to the inspiration derived from our previous work DeepRFT~\cite{XintianMao2023DeepRFT}. In order to utilize kernel-level features more effectively in low training memory, our DeepRFTv2 consists of multiple sub-unets with a reversible strategy \cite{cai2022revcol}.

 
Our main contributions can be summarized as follows:
\begin{itemize}
\item We design an end-to-end deblur model 
by learning kernel-level blur processes in Fourier space. We term the kernel estimation module as Fourier Kernel Estimator (FKE), where kernels are directly convolved on features. Our kernel learning enjoys the benefits of (1) no reblurring supervision and (2) generating global kernels of the same size as the feature. Extensive experiments and visual analyses demonstrate that FKE helps the network to learn the essence of blur degradation.

\item We propose a Decoupled Multi-Scale UNet (DMS-UNet) architecture, designed to utilize kernel-level features more effectively under low training memory constraints, which consists of multiple sub-unets with a reversible strategy. This architecture facilitates bidirectional (top-down and bottom-up) information flow across its hierarchical sub-unets, thereby preventing the premature decoding of these representations. 
\item Extensive experiments show that our method significantly outperforms existing methods on single image deblurring benchmarks and shows potential for handling other kernel-related problems.
\end{itemize}

DeepRFTv2 is an extension of our previous works (DeepRFT~\cite{XintianMao2023DeepRFT} and AdaRevD~\cite{mao2024AdaRevD}) with significant improvements: (1) DeepRFT~\cite{XintianMao2023DeepRFT} designed a Res-FFT-ReLU branch within each ResBlock~\cite{He2016ResNet}, requiring numerous FFT/IFFT operations. Besides, DeepRFT learns kernel-level information, which is then directly added to the spatial feature. This pixel-level learning is hard to interpret how the model's internal representations align with the real kernel properties. By conducting kernel-level learning, DeepRFTv2 performs only a few FFT/IFFT operations. It effectively simulates the physical blur process in the deblurring model, including estimating blur kernels and convolving the kernel with the feature to make the model understand the essence of the blur. (2) AdaRevD~\cite{mao2024AdaRevD} proposed reversible sub-decoders based on existing encoders to improve the decoding capability of the well-trained model and maintain low training memory. Unlike AdaRevD, DeepRFTv2 proposes a decoupled multi-scale network DMS-UNet, which is an end-to-end deblurring model. The reversible strategy is employed to reduce training memory consumption.

\section{Related Work}
\label{sec:related_work}

\subsection{Deep Learning-based Deblurring}
 Previous studies have proposed various CNN-based approaches~\cite{Bahat_2017_ICCV, chakrabarti2016neural, Schuler2016learning, Sun2015learning} to estimate the kernel of the blur image, typically relying on known ground truth blur kernels or those obtained through conventional optimization techniques~\cite{levin2009understanding, whyte2012non}. However, these methods require an accurate kernel for restoration~\cite{Nah2017deep}.
Recently, almost all deblurring methods tend to predict the blur pattern by a network~\cite{Nah2017deep, Tao2018scale, Kupyn2018deblurgan,Kupyn2019deblurgan, Zhang2019DMPHN,Zhang2020deblurring,Zamir2021multi,Cho2021rethinking, XintianMao2023DeepRFT, Chen2021hinet}.
These methods hope to restore the latent sharp image in pixel-level, but this makes it difficult for the network to learn the degradation process caused by the convolution kernel.

To revisit kernel-level learning in the deblur model, UFPNet~\cite{fang2023UFPNet} trains a flow-based model as a kernel estimator to obtain the spatially variant kernel map that is integrated into the deblur model. SegDeblur~\cite{kim2024SegDeblur} applies a segmentation model to obtain the blur segmentation map via kernel estimation, which reflects the characteristics of the image residual error. PGDN~\cite{Fang_PGDN} introduces a parameterized motion kernel estimator using three essential parameters (length, angle, curvature) and builds a dual-branch deblurring network to fully utilize the estimated blur prior information. However, these methods do not utilize their kernels in the convolution manner for the final model optimization. Thus, these methods remain at a pseudo kernel-level. To overcome these problems, we propose a lightweight kernel estimator within the end-to-end model for kernel-level learning. 
\begin{figure}[t!]
\centering

\includegraphics[width=1.0\linewidth]{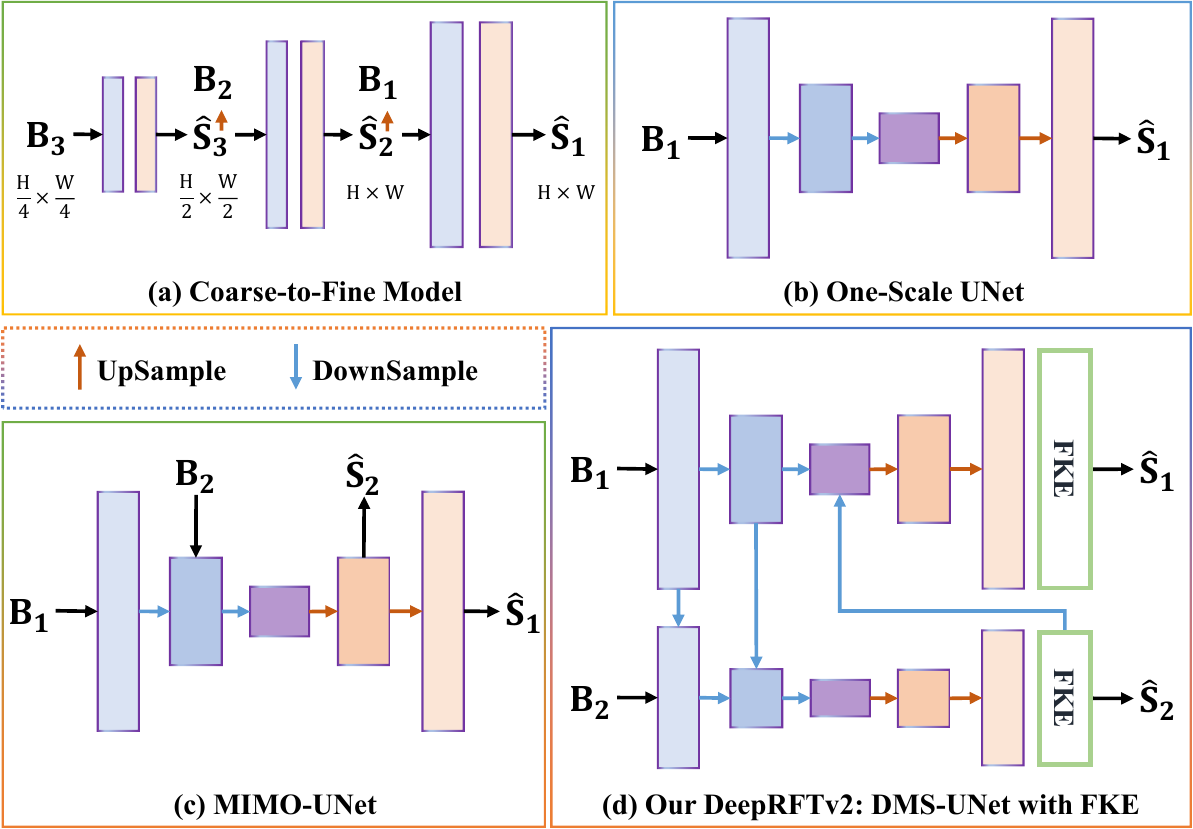}    
\vspace{-1.5em}    
\caption{ (a) Coarse-to-Fine architecture~\cite{Nah2017deep,Tao2018scale}; (b) UNet architecture~\cite{Wang2022uformer,Zamir2021restormer,Chen2022simple}; (c) MIMO-UNet architecture~\cite{Cho2021rethinking,XintianMao2023DeepRFT}; (d) Our DeepRFTv2: DMS-UNet with FKE.  $\mathrm {B}_n$ and $\mathrm {\hat{S}_n}$ represent the blur and restored image at the $n$-th scale. 
                                                  }
                                                  %
\label{fig:DMSUNet}
\vspace{-1.em}
\end{figure}

\subsection{Deep Learning in Frequency Domain}

Many prior works have extracted information from the frequency domain in various vision tasks~\cite{Rippel2015spectral,JunGuo2016DDCN,SunMengdi2020ReductionOJ,Zhong2018joint,Yang2020fda,Zhong2022DetectingCOD,ZequnQin2020FcaNetFC,Rao2021global}. LaMa~\cite{Suvorov2022resolution} applies the Fast Fourier Convolution~\cite{Chi2020fast} to capture a wider received field for image inpainting. For image deblurring, DeepRFT~\cite{XintianMao2023DeepRFT} introduces a Res-FFT-ReLU paradigm for blur pattern recognition. Recently, numerous works have explored frequency information learning in various areas, such as dehazing~\cite{yu2022FSDGN,yu2022FourierUp,cui2025eenet,Hao2025freqdehaze}, super-resolution~\cite{liu2023spectral,wang2023spatial,Arce2024CTSR,guo2024spatially,xiao2024frequency}, enhancement~\cite{yao2024spatial,lv2024Fourier} and other vision tasks~\cite{zhang2025jitter,SFIR,chang2023tsrformer,Yamashita_2024_ACCV,tong2025gsfnet,wu2025joint,yan2025towards,yu2025event,tu2025Fourier}. 
However, these methods require repeated switching between the two domains, which significantly increases computational cost and training memory. In this work, we propose an entire network in Fourier space for global kernel estimation so that the model no longer needs to perform such switch many times.

\begin{figure*}[t]
\begin{center}
    \includegraphics[width=0.9\linewidth]{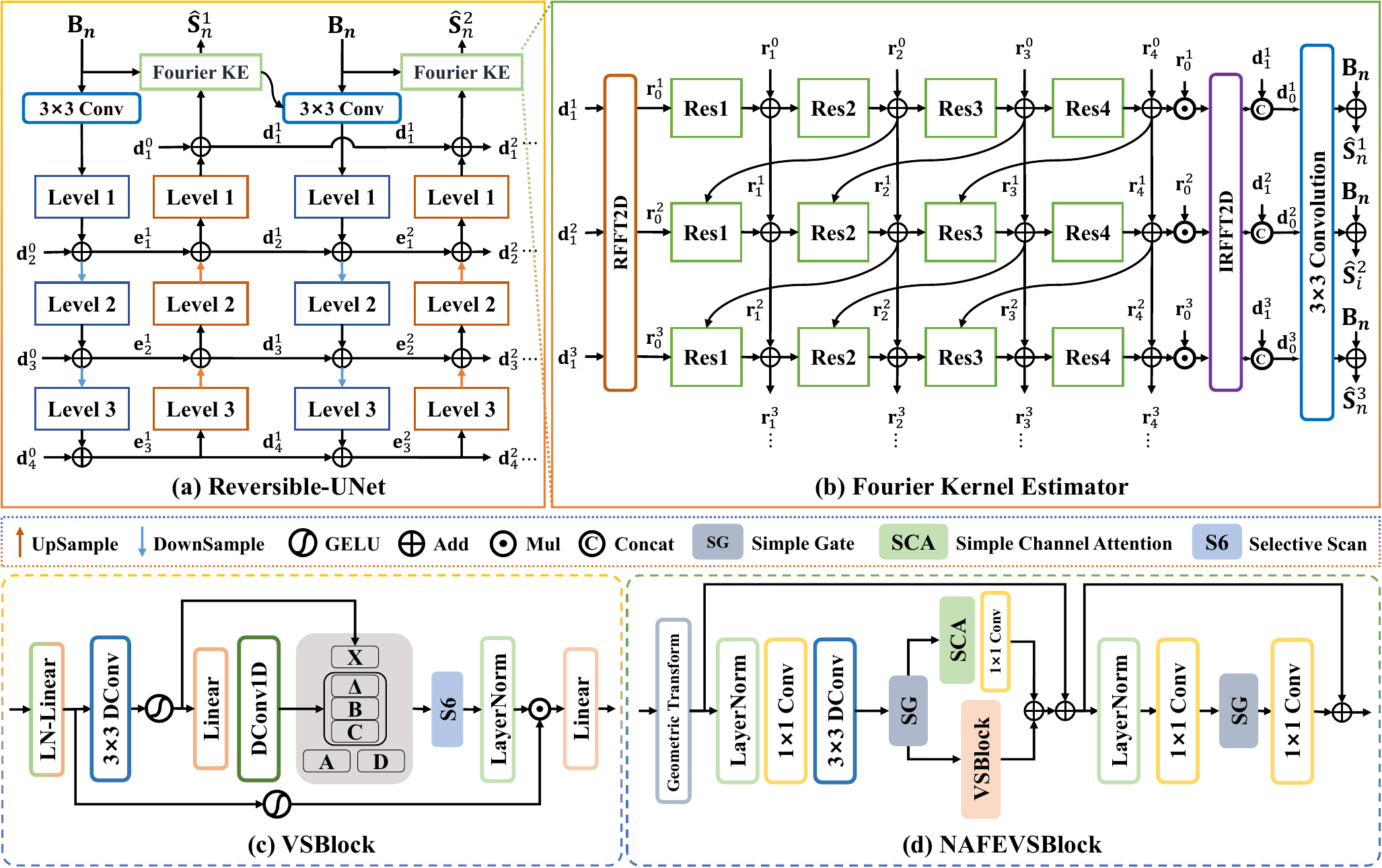}
\end{center}
\vspace{-1.em}
\caption{(a) Reversible-UNet for image restoration with multiple reversible sub-encoders and sub-decoders. Each sub-encoder / decoder is composed of 3 Level module with NAFBlock~\cite{Chen2022simple} and NAFEVSBlock. (b) Fourier Kernel Estimation with multiple reversible sub-resnets. Each sub-resnet is composed of 4 residual modules with VSBlock. (c) VSBlock: Visual Scan Block from EVSSM~\cite{kong2025EVSSM}. (d) NAFEVSBlock: a combination block of NAFBlock~\cite{Chen2022simple} and VSBlock~\cite{kong2025EVSSM} like ALGBlock~\cite{Gao2023ALGNet}.}
\label{fig:rev-unet}
\vspace{-1em}
\end{figure*}

\subsection{Image Restoration Architecture}
Recently, many methods~\cite{Nah2017deep, Zhang2019DMPHN, Zamir2021multi, Zamir2022MIRNetv2, Cho2021rethinking, Liang2021swinir, Tu2022maxim, Chen2022simple, Zamir2021restormer, kong2023fftformer, ren2025reblurring, He_2025_RivuletMLP} applied an end-to-end trained deep neural network for image restoration. In order to achieve better performance, most improvements are made around the model structure or the specific component.


\textbf{Conventional Coarse-to-Fine Model}~\cite{Nah2017deep,Tao2018scale,chen2024NeRD} uses multiple stacks of sub-networks as shown in Fig.~\ref{fig:DMSUNet} (a) to gradually recover the blur image. The network processes input image at multiple scales, with each sub-network handling the image in a different resolution . The conventional coarse-to-fine reconstruction process can be formally described as follows:
\begin{align}
\mathrm {\hat{S}}_n = \mathcal{H}_{{\theta}_n}(\mathrm {B}_n; \mathrm {\hat{S}}^\uparrow _{n+1} ) + \mathrm {B}_n, 
\end{align}
where $\mathcal{H}_{{\theta}_n}$ denotes the sub-network at the $n$-th scale (parameterized by 
${\theta}_n$), $\mathrm {B}_n$ and $\mathrm {\hat{S}_n}$ represent the blur and restored image at the $n$-th scale, respectively, and $\uparrow$ indicates the up-sampling operation. However, coarse-to-fine methods cannot extract the fine details at the beginning.


Compared to conventional coarse-to-fine methods, \textbf{One-Scale UNet}~\cite{UNet} methods~\cite{Chen2022simple,Wang2022uformer,Zamir2021restormer,Tsai2022Stripformer,mao2024Loformer,kong2023fftformer,kong2025EVSSM} exploit multi-scale features extracted from the blur image and predicts latent sharp image in single UNet as shown in Fig.~\ref{fig:DMSUNet} (b). The One-Scale UNet reconstruction process can be described as follows:
\begin{align}
\mathrm {\hat{S}_1}= \mathcal{H}_{\theta}(\mathrm {B_1}) +  \mathrm {B_1}, 
\end{align}
where $\mathcal{H}_{\theta}$ is the One-Scale UNet parameterized by 
${\theta}$. Although the UNet methods are able to extract the fine feature at beginning, it lacks explicit learning of coarse-and-fine image-level features.

To overcome this problem, \textbf{MIMO-UNet}~\cite{Cho2021rethinking} methods~\cite{XintianMao2023DeepRFT,cui2023FSNet,cui2024ConvIR} (Fig.~\ref{fig:DMSUNet} (c)) employ a single UNet with multi-scale shallow feature extraction and latent sharp restoration to achieve more effective deblurring. The encoder is modified to accept multi-scale input images, while its decoder is adapted to produce multi-scale latent sharp outputs during decoding, enabling more effective coarse-to-fine deblurring. The MIMO-UNet reconstruction process can be formally described as follows:
\begin{align}
\left \{  \mathrm {\hat{S}_1}, ..., \mathrm {\hat{S}}_N \right \}= \mathcal{H}_{\theta}(\mathrm {B_1}, ..., \mathrm {B}_N) + \left \{ \mathrm {B_1}, ..., \mathrm {B}_N \right \}, 
\end{align}
where $\mathcal{H}_{\theta}$ is the MIMO-UNet parameterized by 
${\theta}$.

However, introducing low-level information into the degradation representation learning process directly may interfere with the encoding process. Meanwhile, the decoder may decode the degradation representation to blur patterns prematurely. To solve this problem, we decouple the MIMO-UNet to DMS-UNet with multiple hierarchical sub-unets, which facilitates bidirectional multi-scale information flow.


\section{Methods}
\label{sec:method}
The schematic of the proposed DeepRFTv2 is shown in Fig.~\ref{fig:DMSUNet}(d), which consists of the DMS-UNet architecture and the FKE module. Each UNet structure in our DeepRFTv2 is referred to as a Reversible-UNet, composed of multiple reversible sub-encoders, sub-decoders and FKE modules, as shown in Fig.~\ref{fig:rev-unet} (a).
We first provide the details of \textbf{FKE} in section~\ref{sec:FKE}, which is a core component of DeepRFTv2.
Then, \textbf{DMS-UNet} will be introduced in section~\ref{sec:DMS-UNet}, which is the feature extraction model of DeepRFTv2.
Finally, the basic modules of DeepRFTv2 will be described in section~\ref{sec:Basic-Modules}. 

\subsection{Fourier Kernel Estimator}
\label{sec:FKE}
As delineated in Eq. \ref{eq:blur}, and under the assumption of negligible additive noise, recovering the blur kernel in the spatial domain requires learning a deconvolution process. However, this is inherently challenging for neural architectures, as their fundamental operations, such as addition, multiplication, and activation functions, are ill-suited to directly model the deconvolution process.


Unlike the spatial domain, Fourier space presents a more tractable framework for blur kernel estimation. DeepRFT~\cite{XintianMao2023DeepRFT} has revealed that the activation in Fourier space of the blur image contains \textbf{kernel pattern information}. As illustrated in Fig.~\ref{fig:fft_relu}, after applying FFT-ReLU-IFFT, subtracting half of the blur images and cyclic shift with the halves of height and width, a phenomenon emerges: the results exhibit a notable similarity to the blur mode of the images. As Eq. \ref{eq:decblur} shows, the complex deconvolution problem can be transformed into a simpler division operation:
\begin{align}
\mathcal{F}(\mathrm{k}) = \overline{\mathcal{F}(\mathrm{S})}\odot\mathcal{F}(\mathrm{B}) / (\overline{\mathcal{F}(\mathrm{S})}\odot\mathcal{F}(\mathrm{S})),
\label{eq:decblur}
\end{align}
where $\overline{\mathcal{F}(\mathrm{S})}$ is the conjugate of $\mathcal{F}(\mathrm{S})$. This conversion makes it easier for kernel estimation. 
In summary, these analytical and empirical findings support the feasibility of kernel estimation in Fourier space.

We propose Fourier Kernel Estimator (FKE) to estimate the blur kernel and learn the convolution process defined in Eq.~\ref{eq:blur}. Compared to images, the multi-channel features extracted by the network provide richer semantic and structural information, which are more suitable for global convolution kernels. As can be seen in Fig.~\ref{fig:rev-unet} (a) and (b), we put our FKE module in the tail part of each sub-unet which takes the feature $\mathrm {d}_1^j$ as input and outputs the global kernels for convolution. The overall process can be formally described as follows:
\begin{align}
\hat{\mathrm{S}} &= \mathcal{K}_{{\delta}}(\mathcal{H}_{{\theta}}(\mathrm{B})) + \mathrm{B},
\label{eq:DeepRFTv2}
\end{align}
where $\mathcal{K}_{{\delta}}$ and $\mathcal{H}_{{\theta}}$ denote the FKE module and the feature extraction module parameterized by ${\delta}$ and ${\theta}$, respectively.

As shown in Fig.~\ref{fig:rev-unet} (b), the FKE is composed of several residual blocks that take the frequency feature $\mathrm{r}_0^j=\mathcal{F}(\mathrm{d}_1^j)\in\mathbb{C}^{\rm{H}\times\rm{W}\times\rm{C}}$ as the input and outputs the kernels in Fourier space. Since convolving the image with the estimated kernel $\mathrm{k}\in\mathbb{C}^{\rm{H}\times\rm{W}\times\rm{C}}$ is always affected by other factors such as additive noise, we replace the image with the features that have been fully extracted in order to gain more freedom and stability. $\mathrm{r}_0^j$ is then multiplied by the kernels in Fourier space, which is equal to convolution in the spatial domain. The process of $\mathcal{K}_{{\delta}}(\cdot)$ can be formally described as follows:
\begin{align}
\hat{\mathrm{P}}=\mathbf{Conv}(\mathcal{F}^{-1}(\mathcal{R}_{\theta}^j(\mathrm{r}_0^j)\odot\mathrm{r}_0^j)~\textcircled{c}~\mathrm{d}_1^j),
\end{align}
where $\mathbf{Conv}$ denotes a $3\times3$ convolution for dimension reduction, taking the concatenation (denoted by $\textcircled{c}$) of $\mathcal{F}^{-1}(\mathrm{r}_4^j)\odot\mathrm{r}_0^j$ and $\mathrm{d}_1^j$ as input. Besides, $\mathcal{R}_{\theta}^j$ represents the $j$-th sub-resnet parameterized by ${\theta^j}$, $\mathrm{r}_0^j$ is the feature in Fourier space, and $\mathrm {\hat{P}}$ is the predicted blur pattern.
\begin{figure*}[t!]

    \begin{center}
        \includegraphics[width=0.9\linewidth]{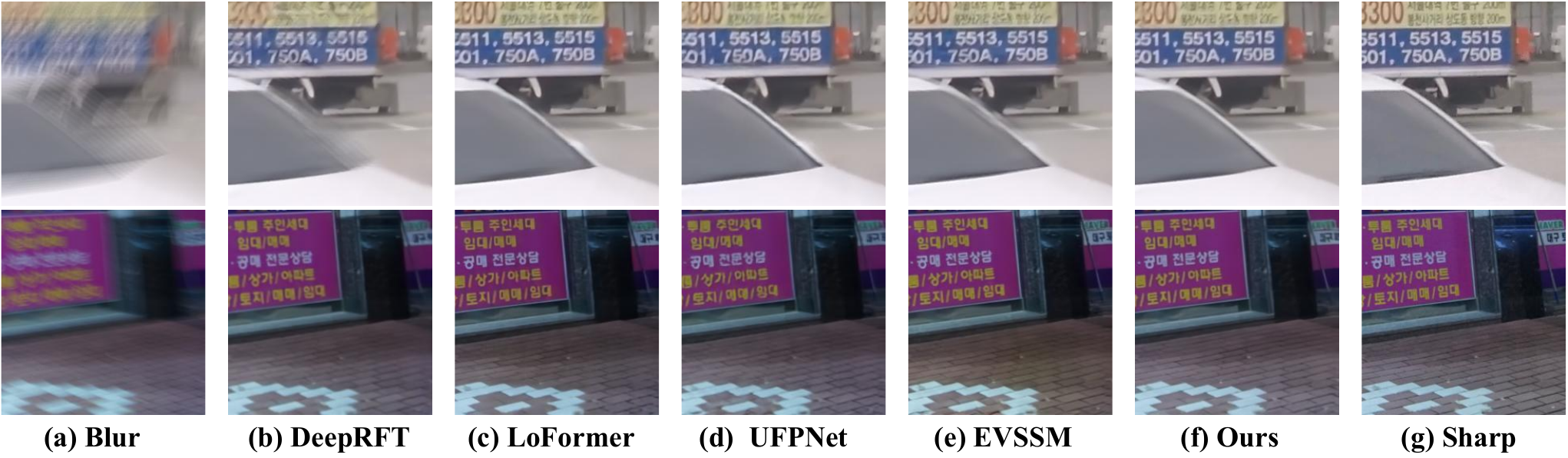}
    \end{center}
    \vspace{-1em}
        \caption{Visual comparison of single image motion deblur approaches on GoPro~\cite{Nah2017deep} and RealBlur-J~\cite{Rim2020real}.}
        \label{fig:gopro_realblur}
        \vspace{-1.em}
\end{figure*}

Based on the above process, our FKE is designed to directly estimate physically plausible kernels (Fig.~\ref{fig:feature_kernel}) from the frequency spectrum by a light-weight network. We choose to conduct kernel estimation using the features obtained through the feature extraction module, following the forward kernel prior-assisted methods \cite{fang2023UFPNet,Fang_PGDN}. Unlike them, our method does not need additional kernel supervision. The efficacy of FKE stems from the convolution theorem: by performing convolution via kernel-feature multiplication in Fourier space, it intrinsically learns the essence of the blur. This direct modeling empowers the network with a more accurate representation of the degradation process, leading to superior deblurring performance.

\subsection{Decoupled Multi-Scale UNet}
\label{sec:DMS-UNet}
As shown in Fig.~\ref{fig:DMSUNet} (d), DMS-UNet consists of \textbf{decoupled fine-to-coarse encoders} and \textbf{decoupled coarse-to-fine decoders}, which allows top-to-bottom and bottom-to-top information flow in hierarchical \textbf{reversible sub-unets}. Due to the decoupled architecture, each sub-unet can maintain a certain degree of independence during the feature encoding and decoding process for image deblurring.

\textbf{Decoupled Fine-to-Coarse Encoder} extracts features by multi-scale progressive encoders from multi-scale input images. The decoupled top-to-bottom information flow enables each encoder to progressively refine features from fine to coarse without interference, ensuring the independent learning of multi-scale feature representations. The decoupled fine-to-coarse encoder process can be formally described as follows:
\begin{align}
\left \{  \mathrm {e}_{i,n}^1, ..., \mathrm {e}_{i,n}^j \right \}= \mathcal{E}_{\theta_{n}^j}(\mathrm {B}_n; \mathrm {e}_{i,n-1}^1, ..., \mathrm {e}_{i,n-1}^j), 
\end{align}
where $\mathcal{E}_{\theta_n^j}$ and $\mathrm {e}_{i,n}^j$ denote the $j$-th encoder at the $n$-th scale (parameterized by ${\theta}$) and its output feature at the $i$-th level.

\textbf{Decoupled Coarse-to-Fine Decoder} decodes the degradation representation to the blur pattern from small scale to large scale by multi-scale progressive decoders. Through the decoupled bottom-to-top information flow, each decoder is able to adequately decode the degradation representation to the blur pattern. The decoupled coarse-to-fine decoder process can be formally described as follows:
\begin{align}
\left \{  \mathrm {d}_{i,n}^1; ...; \mathrm{d}_{i,n}^j \right \}&= \mathcal{D}_{\theta_{n}^j}(\mathrm {e}_{i,n}^1, ..., \mathrm {e}_{i,n}^j; \mathrm {d}_{i,n+1}^0, ..., \mathrm {d}_{i,n+1}^j), \\
\hat{\mathrm{S}}_n^j&=\mathcal{T}_{\theta_{n}^j}(\mathrm{d_1^j})+{\mathrm{B}_n},
\end{align}
where $\mathcal{D}_{\theta_n^j}$ and $\mathcal{T}_{\theta_n^j}$ denote the $j$-th sub-decoder and its tail module at the $n$-th scale (parameterized by ${\theta_{n}^j}$). $\mathrm {d}_{i,n}^j$ and $\mathrm {d}_{i,n}^0$ is the output feature of $\mathcal{D}_{\theta_n^j}$ and its FKE module. $\mathrm{B}_n$ and $\mathrm {\hat{S}}_n^j$ are the blur input and restored output, respectively.

\subsection{Basic Modules}
\label{sec:Basic-Modules}

\subsubsection{Reversible UNet for \textbf{DMS-UNet}}
DeepRFTv2 is trained by reversible strategy~\cite{gomez2017reversible,jacobsen2018irevnet,mangalam2022revViT,chiley2023revBiFPN,cai2022revcol} for lower GPU memory. Fig.~\ref{fig:rev-unet} (a) illustrates the architecture of the reversible sub-unets. Each sub-unet comprises a single sub-encoder and a single sub-decoder. Formally, the forward and inverse computations of the sub-encoder are defined as follows:
\begin{align}
\text{Forward: }\mathrm{e}_i^j=\begin{cases}
     ~\mathcal{L}_i^j(\mathrm{e}_{i-1}^{j})^\downarrow ~+~ \alpha \mathrm{d}_{i+1}^{j-1}&i > 1 
    \\
    ~\mathcal{L}_i^j(\mathbf{Conv}(\mathrm{B}_{n})) ~+~ \alpha \mathrm{d}_{i+1}^{j-1}&i = 1  
    \end{cases} \\
\text{Inverse: }\mathrm{d}_{i+1}^{j-1}=\begin{cases}
     \alpha^{-1}(\mathrm{e}_i^j-\mathcal{L}_i^j(\mathrm{e}_{i-1}^{j})) ^\downarrow&i > 1  
    \\
    \alpha^{-1}(\mathrm{e}_i^j-\mathcal{L}_i^j(\mathbf{Conv}(\mathrm{B}_{n})))&i = 1  
    \end{cases}
\end{align} 
where $\mathcal{L}_i^j$ and $\mathbf{e}_i^j$ denote the $i$-th Level module and its output feature in the $j$-th sub-encoder, respectively, while $^\downarrow$ and $\mathbf{Conv}$ represent downsampling operation and $3\times3$ convolution feature extraction.

The forward and inverse computations of the sub-decoders are defined as follows :
\begin{align}
\text{Forward: }\mathrm{d}_i^j~=\begin{cases}
     ~\mathcal{L}_i^j(\mathrm{d}_{i+1}^{j})^\uparrow ~+~ \alpha \mathrm{e}_{i-1}^{j} &i > 1  
    \\
    ~\mathcal{L}_i^j(\mathrm{d}_{i+1}^{j}) ~+~ \alpha \mathrm{d}_{i}^{j-1} &i = 1  
    \end{cases} \\ 
\text{Inverse: }\begin{cases}
     \mathrm{e}_{i-1}^{j}=\alpha^{-1}(\mathrm{d}_i^j-\mathcal{L}_i^j(\mathrm{d}_{i+1}^{j})^\uparrow) &i > 1  \\
    \mathrm{d}_{i}^{j-1}=\alpha^{-1}(\mathrm{d}_i^j-\mathcal{L}_i^j(\mathrm{d}_{i+1}^{j})) &i = 1  
    \end{cases}
\end{align} 
where $\mathcal{L}_i^j$ and $\mathrm{d}_i^j$ denote the $i$-th Level module and its output feature of the $j$-th sub-decoder, respectively, while $^\uparrow$ represents the upsampling operation.

\subsubsection{Reversible ResNet for \textbf{FKE}}
As shown in Fig.~\ref{fig:rev-unet} (b), we build our kernel estimator model by multiple reversible sub-resnet. Formally, the forward and inverse computations of the sub-resnet are:
\begin{align}
\text{Forward: }~\mathrm{r}_i^j=\begin{cases}
     \mathcal{R}_i^j(\mathrm{r}_{i-1}^j, \mathrm{r}_{i-1}^{j+1}) + \alpha \mathrm{r}_i^{j-1}~~~~~&i > 1  
    \\
    \mathcal{R}_i^j(\mathrm{r}_{i-1}^j) + \alpha \mathrm{r}_i^{j-1}~~~~~&i = 1  
    \end{cases}\\
\text{Inverse: }\mathrm{r}_i^{j-1}=\begin{cases}
     \alpha^{-1}(\mathrm{r}_i^j-\mathcal{R}_i^j(\mathrm{r}_{i-1}^j, \mathrm{r}_{i+1}^{j-1}))~&i > 1  
    \\
    \alpha^{-1}(\mathrm{r}_i^j-\mathcal{R}_i^j(\mathrm{r}_{i-1}^j))~&i = 1  
    \end{cases}
\end{align}
where $\mathcal{R}_i^j$ and $\mathrm{r}_i^j$ denote the $i$-th residual module and its output feature of the $j$-th ResNet, respectively, while $\alpha$ is a learnable scaling parameter.

\subsubsection{Basic Block for \textbf{DeepRFTv2}} 
Based on the strength of the state space model (SSM)~\cite{gu2021LSSL,2022gu_s4,gu2023mamba} in modeling long-range dependencies with linear complexity, a growing number of studies~\cite{guo2025MambaIR,guo2025mambairv2,shi2025vmambair,sun2024hybrid,liFouriermamba,Gao2023ALGNet,Zhao_2025_CVPR,kong2025EVSSM,Fang_PGDN,Liu_2025_XYScanNet} have successfully applied them to image restoration. Following this trend, we adopt established and effective modules as the foundational units for DMS-UNet (Level module) and FKE (Res module). As illustrated in Fig.~\ref{fig:rev-unet} (c) and (d), VSBlock~\cite{kong2025EVSSM}, NAFBlock~\cite{Chen2022simple} and NAFEVSBlock are the basic blocks of DeepRFTv2.
\begin{table}[t]
\footnotesize
\begin{center}
\caption{Comparisons on GoPro~\cite{Nah2017deep} and HIDE~\cite{Shen2019human}. GPU memory is calculated by training a single patch
$256\times256$.}
\label{tab:trained on GoPro}
\vspace{-0.5em}
\renewcommand\arraystretch{0.8}
\setlength{\tabcolsep}{1.9pt}
\begin{tabular}{l c | c | cc }
\toprule[0.15em]
 & \textbf{GoPro} & \textbf{HIDE} &  \textbf{FLOPs} & \textbf{Memory} \\
 \textbf{Method} & PSNR~\colorbox{color4}{SSIM} & PSNR~\colorbox{color4}{SSIM} &  (G) & (MB)\\
\midrule[0.15em]
DeepDeblur~\cite{Nah2017deep}  & 29.08 \colorbox{color4}{0.914} & 25.73 \colorbox{color4}{0.874}  & 336 & - \\
MPRNet~\cite{Zamir2021multi} & 32.66 \colorbox{color4}{0.959} & {30.96} \colorbox{color4}{0.939} & 777 & 6,294 \\
DeepRFT~\cite{XintianMao2023DeepRFT} & 33.52 \colorbox{color4}{0.965} & 31.66 \colorbox{color4}{0.946} & 187 & 3,569 \\
NAFNet~\cite{Chen2022simple} & 33.69 \colorbox{color4}{0.967} &31.32  \colorbox{color4}{0.943} & 64 & 2,972  \\
{Restormer}~\cite{Zamir2021restormer} & {32.92} \colorbox{color4}{{0.961}} & {31.22} \colorbox{color4}{{0.942}}  & 141 & 12,520\\
{MRLPFNet}~\cite{dong2023MRLPFNet} & 34.01 \colorbox{color4}{0.968} & 31.63 \colorbox{color4}{0.947} & 129 & - \\
ALGNet~\cite{Gao2023ALGNet} & 34.05 \colorbox{color4}{0.969} & 31.68 \colorbox{color4}{0.952} & - & - \\
{LoFormer~\cite{mao2024Loformer}} & 34.09 \colorbox{color4}{0.969} & 31.86 \colorbox{color4}{0.949} & 126 & 6,212\\
FFTformer~\cite{kong2023fftformer} & 34.21 \colorbox{color4}{0.968} & 31.62 \colorbox{color4}{0.946} & 132 & 19,800 \\
MISCFilter~\cite{liu2024miscfilter} & 34.10 \colorbox{color4}{0.969} & 31.66 \colorbox{color4}{0.946} & - & - \\
MDT~\cite{chen2025MDT} & 34.26 \colorbox{color4}{0.969} & 31.84 \colorbox{color4}{0.948} & 113  & -   \\
UFPNet~\cite{fang2023UFPNet} & 34.06 \colorbox{color4}{0.968} & 31.74 
\colorbox{color4}{0.947} & 243 & 8,164 \\
{PGDN~\cite{Fang_PGDN}} & 34.17 \colorbox{color4}{0.950} & - & - & -  \\

\arrayrulecolor{black!30}\midrule
DeepRFTv2-B & \textbf{34.52} \colorbox{color4}{0.971} & \textbf{32.22} \colorbox{color4}{0.953} & 120 & 3,439 \\
\arrayrulecolor{black}\bottomrule[0.15em]
\end{tabular}
\end{center}
\vspace{-2.em}
\end{table}

\section{Experiment}
\label{sec:experiments}

\subsection{Experimental Setup}
\label{sec:dataset-detail}
\textbf{Dataset} We evaluate our method on various motion-deblur, defocus-deblur and super-resolution datasets and report five groups of results:

\noindent\textcolor{violet}{$\mathcal{A.}$}~train on GoPro, test on GoPro~\cite{Nah2017deep} / HIDE~\cite{Shen2019human};

\noindent\textcolor{violet}{$\mathcal{B.}$}~train and test on RealBlur-R / J~\cite{Rim2020real} respectively;

\noindent\textcolor{violet}{$\mathcal{C.}$}~train and test on RED-val-300~\cite{Nah2021ntire};

\noindent\textcolor{violet}{$\mathcal{D.}$}~train and test on DPDD~\cite{Abuolaim2020DPDD};

\noindent\textcolor{violet}{$\mathcal{E.}$}~train on DF2K~\cite{lim2017edsr,timofte2017ntire} and test on Set5~\cite{Bevilacqua2012set5}, Set14~\cite{Zeyde_2012_set14}, BSD100~\cite{Martin_2002_BSD100},
Urban100~\cite{Huang_2015_Urban100}, Manga109~\cite{Matsui_2016_Manga109}.

\noindent\textbf{Configuration} We build two models for comparison with other methods: DeepRFTv2-S, DeepRFTv2-B. All of them are constructed by two DMS-UNets, which consists of two scales. The numbers of basic blocks in the sub-encoder / sub-decoder from Level 1 to Level 3 are [1, 4, 8], and the respective channel numbers are [48, 96, 192].  The basic block of DeepRFTv2-S is NAFBlock~\cite{Chen2022simple}. Compared to DeepRFTv2-S, NAFEVSBlock is applied for Level 2 and 3 of the first scale in DeepRFTv2-B. The tail part of each sub-decoder is FKE. VSBlock~\cite{kong2025EVSSM} with skip addition is the basic block of FKE. The numbers of basic blocks from Res-1 to Res-4 are [1, 1, 1, 1], and the respective channel numbers are [48, 48, 48, 48]. 

DeepRFTv2 is supervised exclusively by deblurring loss, defined as the summation of L1 loss ($\ell_{1}$) and Frequency Reconstruction loss ($\ell_{freq}$): $\ell_{total}=\ell_{1}+0.01\ell_{freq}$.

We adopt the training strategy from NAFNet~\cite{Chen2022simple} unless otherwise specified. The network training hyperparameters (and the default values) are optimizer Adam~\cite{kingma2014adam} ($\beta_1=0.9$, $\beta_2=0.9$, weight decay 1$\times$10$^{-3}$), initial learning rate (1$\times$10$^{-3}$). The learning rate is steadily decreased to 1$\times$10$^{-7}$ with the cosine annealing schedule. Due to variations in the size of the image in different datasets, we adaptively designed slightly divergent training strategies:

\noindent\textcolor{violet}{$\mathcal{A.}$} For GoPro~\cite{Nah2017deep}, DeepRFTv2-B is progressively trained with patch size (256$\times$256, 512$\times$512) and batch size (16, 8) for (450K, 150K) iterations;

\noindent\textcolor{violet}{$\mathcal{B.}$} For RealBlur~\cite{Rim2020real}, due to the slight differences in image size, we uniformly zero-padded them to the same size (776$\times$680) for training, and during inference, the padding parts would be cut off. DeepRFTv2-B is progressively trained with patch size (512$\times$512, 776$\times$680) and batch size (16, 8) for (50K, 100K) iterations;

\noindent\textcolor{violet}{$\mathcal{C.}$} For REDS-val-300~\cite{Nah2021ntire}, DeepRFT-B is trained with patch size 512$\times$512 and batch size 8 for 600K iterations;

\noindent\textcolor{violet}{$\mathcal{D.}$} For DPDD~\cite{Abuolaim2020DPDD}, DeepRFT-S is trained with patch size 640$\times$640 and batch size 8 for 200K iterations;

\noindent\textcolor{violet}{$\mathcal{E.}$} For lightweight SR, FKE is placed at the end of each residual module in ATD~\cite{zhang2024ATD}. The training strategy is copied from ATD's training strategy.

\begin{table}[t]
\footnotesize
\begin{center}
\caption{Comparisons on RealBlur~\cite{Rim2020real}. Rows in \colorbox{gray!10}{gray} means computing without aligning the deblurring result to their ground truth sharp image.}
\vspace{-0.5em}
\label{tab:trained on RealBlur}
\renewcommand\arraystretch{0.7}
\setlength{\tabcolsep}{1.9pt}
\begin{tabular}{l c | c | c  }
\toprule[0.15em]
 & \textbf{RealBlur-R} & \textbf{RealBlur-J} & \textbf{Average}  \\
 \textbf{Method} & PSNR~\colorbox{color4}{SSIM} & PSNR~\colorbox{color4}{SSIM} & PSNR~\colorbox{color4}{SSIM}  \\
 \midrule
SegDeblur~\cite{kim2024SegDeblur}  & {40.79} \colorbox{color4}{0.976} & {33.51} \colorbox{color4}{0.938} & 37.15  \colorbox{color4}{0.957} \\
AdaRevD~\cite{mao2024AdaRevD}  & {41.09} \colorbox{color4}{{0.978}} & {33.84} \colorbox{color4}{{0.943}}& {37.47} \colorbox{color4}{{0.961}} \\
FFTformer~\cite{kong2023fftformer} & 40.11 \colorbox{color4}{0.975} & 32.62 \colorbox{color4}{0.933} & 36.37 \colorbox{color4}{0.954}   \\
LoFormer~\cite{mao2024Loformer}  & {40.50} \colorbox{color4}{0.975} & {32.88} \colorbox{color4}{0.935} & 36.69  \colorbox{color4}{0.955} \\
UFPNet~\cite{fang2023UFPNet} & 40.61 \colorbox{color4}{0.974} & {33.35} \colorbox{color4}{0.934} & 36.98 \colorbox{color4}{0.954}  \\
EVSSM~\cite{kong2025EVSSM} & {41.27} \colorbox{color4}{0.978} & 34.34 \colorbox{color4}{0.946}  & 37.81 \colorbox{color4}{0.962}  \\
\arrayrulecolor{black!30}\midrule
DeepRFTv2-B  & \textbf{41.39} \colorbox{color4}{0.977} & \textbf{34.46} \colorbox{color4}{{0.945}}& \textbf{37.93} \colorbox{color4}{0.961}\\
\arrayrulecolor{black!60}\midrule
\rowcolor{gray!10} FFTformer~\cite{kong2023fftformer} & 37.32 \colorbox{gray!20}{0.966} & {29.97} \colorbox{gray!20}{0.922} & 33.65 \colorbox{gray!20}{0.944}  \\
\rowcolor{gray!10} LoFormer~\cite{mao2024Loformer} & {37.67} \colorbox{gray!20}{{0.968}} & {30.04} \colorbox{gray!20}{{0.924}} &33.86 \colorbox{gray!20}{0.946} \\
\rowcolor{gray!10} UFPNet~\cite{fang2023UFPNet} & 38.14 \colorbox{gray!20}{0.971} & {30.82} \colorbox{gray!20}{0.929} & 34.48 \colorbox{gray!20}{0.950}  \\
\rowcolor{gray!10} EVSSM~\cite{kong2025EVSSM} & 37.82 \colorbox{gray!20}{0.970} & {30.68} \colorbox{gray!20}{0.929} & 34.25 \colorbox{gray!20}{0.950}  \\
\arrayrulecolor{black!30}\midrule
\rowcolor{gray!10} DeepRFTv2-B & \textbf{38.26} \colorbox{gray!20}{{0.971}} & \textbf{31.10} \colorbox{gray!20}{{0.933}} & \textbf{34.68} \colorbox{gray!20}{0.952}  \\
\arrayrulecolor{black}\bottomrule[0.15em]
  \end{tabular}
\end{center}
\vspace{-0.9em}
\end{table}

\begin{table}[t]
\footnotesize
\begin{center}
\caption{Comparison on the REDS-val-300~\cite{Nah2021ntire} of NTIRE 2021 Image Deblurring Challenge Track 2 JPEG artifacts.}
\vspace{-0.5em}
\label{tab:REDS-val-300}
\begin{tabular}{lcccc}
\toprule[0.15em]
\textbf{Model}& \textbf{PSNR}& \textbf{SSIM} & \textbf{FLOPs} & \textbf{Params} \\
\midrule[0.09em]
MPRNet~\cite{Zamir2021multi} & 28.79 & 0.811 & 777 & 20.1 \\
HINet~\cite{Chen2021hinet} & 28.83 & 0.862 & 171 & 88.7 \\
MAXIM~\cite{Tu2022maxim} & 28.93 & 0.865  & 339 & 22.2 \\
NAFNet~\cite{Chen2022simple} & 29.09 & 0.867  & 64 & 65.0 \\
LoFormer~\cite{mao2024Loformer} & {29.20} & {0.869} & 73 & 27.9 \\
\arrayrulecolor{black}\midrule
DeepRFTv2-B & \textbf{29.42} & \textbf{0.872} & 120 & 30.1 \\
\bottomrule[0.15em]
\end{tabular} 
\end{center}
\vspace{-1.5em}
\end{table}

 \begin{table*}[!t]
    \footnotesize
        \centering
        \captionof{table}{Comparisons with other \textbf{Single Image Defocus Deblurring} methods on the DPDD~\cite{Abuolaim2020DPDD}.}
        \vspace{-0.5em}
        \setlength{\tabcolsep}{1.9pt}
    \begin{tabular}{l|cccc|cccc|cccc}
    \toprule[0.15em] 
    \multicolumn{1}{c|}{} & \multicolumn{4}{c|}{Indoor Scenes}  & \multicolumn{4}{c|}{Outdoor Scenes} & \multicolumn{4}{c}{Combined}
    \\
   ~Methods~ & ~~PSNR$\uparrow$~~ & ~~SSIM$\uparrow$~~ &~~MAE$\downarrow$~~  &~~LPIPS$\downarrow$~~ & ~~PSNR$\uparrow$~~ & ~~SSIM$\uparrow$~~  &~~MAE$\downarrow$~~  &~~LPIPS$\downarrow$~~  &  ~~PSNR$\uparrow$~~ &  ~~SSIM$\uparrow$~~ &~~MAE$\downarrow$~~  &~~LPIPS$\downarrow$~~ 
    \\
    \midrule
    ~EBDB~\cite{Karaali2018edge}& 25.77 &0.772 &0.040 &0.297 &21.25 &0.599 &0.058 &0.373 &23.45 &0.683 &0.049 &0.336
    \\
~DMENet~\cite{Lee2019deep}~~  &25.50 &0.788 &0.038 &0.298 &21.43 &0.644 &0.063 &0.397 &23.41 &0.714 &0.051 &0.349
\\
~JNB~\cite{Shi2015just} &26.73 &0.828 &0.031 &0.273 &21.10 &0.608 &0.064 &0.355 &23.84 &0.715 &0.048 &0.315
\\
~DPDNet~\cite{Abuolaim2020DPDD} & 26.54 &0.816 &0.031 &0.239 &22.25 &0.682 &0.056 &0.313 &24.34 &0.747 &0.044& 0.277
\\
~KPAC~\cite{Son_2021_KPAC}& 27.97 &0.852 &0.026 &0.182 &22.62 &0.701 &0.053 &0.269 &25.22 &0.774 &0.040 &0.227
\\
~IFAN~\cite{Lee2021iterative}& 28.11 &0.861 &0.026 &0.179 &22.76 &0.720 &0.052 &0.254 &25.37 &0.789 &0.039 &0.217
\\
~ConvIR~\cite{cui2024ConvIR} &\underline{29.37} &0.887 &\textbf{0.023} &\underline{0.143} &{23.51} &\underline{0.757} &{0.049} &\underline{0.203} &{26.36} &{0.820} &\textbf{0.036} &\underline{0.174}
\\
~{ALGNet~\cite{Gao2023ALGNet}}&\underline{29.37}&\underline{0.888}	&\textbf{0.023}	&{0.147}	&\underline{23.68}	&{0.755}	&\textbf{0.048}	&{0.223}	&\underline{26.45}	&\underline{0.821}	&\textbf{0.036}	&{0.186}
\\
~FDIKP~\cite{zhang2025FDIKP} &- &- &-	&- &- &- &- &- &26.42	&0.813	&-	&{0.185}\\
\arrayrulecolor{black!30}\midrule
~\textbf{DeepRFTv2-S}&\textbf{29.50}	&\textbf{0.890}	&\textbf{0.023}	&\textbf{0.134}	&\textbf{23.73}	&\textbf{0.766}	&\textbf{0.048}	&\textbf{0.192}	&\textbf{26.54}	&\textbf{0.827}	&\textbf{0.036}	&\textbf{0.164}\\

    \arrayrulecolor{black}\bottomrule[0.15em]
    \end{tabular}
   
        
        \label{tab:defocus}
        \end{table*}

\begin{table*}

\caption{Comparison on \textbf{lightweight Super-Resolution}. The best and second best results are colored with \color{red}{red} \color{black}{and} \color{blue}{blue}. }
\vspace{-3mm}
\label{tab: results image sr2}
  \begin{center}

  \begin{tabular}{p{3.cm}|c|c|cc|cc|cc|cc|cc}
    \toprule
    \multirow{2}{*}{\textbf{Method}} & \multirow{2}{*}{\textbf{Scale}} & \multirow{2}{*}{\textbf{Params}} & \multicolumn{2}{c|}{\textbf{Set5}} & \multicolumn{2}{c|}{\textbf{Set14}} & \multicolumn{2}{c|}{\textbf{BSD100}} & \multicolumn{2}{c|}{\textbf{Urban100}} & \multicolumn{2}{c}{\textbf{Manga109}} \\

    & & & PSNR & SSIM & PSNR & SSIM & PSNR & SSIM & PSNR & SSIM & PSNR & SSIM   \\

    \midrule
    CARN~\cite{Ahn_2018_carn}               & $\times$2 & 1,592K & 37.76 & 0.9590 & 33.52 & 0.9166 & 32.09 & 0.8978 & 31.92 & 0.9256 & 38.36 & 0.9765 \\
    IMDN~\cite{Hui_2019_imdn}               & $\times$2 & 694K   & 38.00 & 0.9605 & 33.63 & 0.9177 & 32.19 & 0.8996 & 32.17 & 0.9283 & 38.88 & 0.9774 \\
    LAPAR-A~\cite{Li_2020_lapar}            & $\times$2 & 548K   & 38.01 & 0.9605 & 33.62 & 0.9183 & 32.19 & 0.8999 & 32.10 & 0.9283 & 38.67 & 0.9772 \\
    LatticeNet~\cite{Luo_2020_latticenet}   & $\times$2 & 756K   & 38.15 & 0.9610 & 33.78 & 0.9193 & 32.25 & 0.9005 & 32.43 & 0.9302 & -     & -      \\
    SwinIR-light~\cite{Liang2021swinir}     & $\times$2 & 910K   & 38.14 & 0.9611 & 33.86 & 0.9206 & 32.31 & 0.9012 & 32.76 & 0.9340 & 39.12 & 0.9783 \\
    ELAN~\cite{zhang2022efficient}          & $\times$2 & 582K   & 38.17 & 0.9611 & 33.94 & 0.9207 & 32.30 & 0.9012 & 32.76 & 0.9340 & 39.11 & 0.9782 \\
    SwinIR-NG~\cite{Choi_2022_swinirng}     & $\times$2 & 1181K  & 38.17 & 0.9612 & 33.94 & 0.9205 & 32.31 & 0.9013 & 32.78 & 0.9340 & 39.20 & 0.9781 \\
    OmniSR~\cite{omni_sr}                   & $\times$2 & 772K   & {38.22} & {0.9613} & {33.98} & {0.9210} & {32.36} & {0.9020} & {33.05} & {0.9363} & {39.28} & {0.9784} \\
    MambaIRv2-light~\cite{guo2025mambairv2}       & $\times$2 & 774K   & {38.26} & {0.9615} & {34.09} & \sotab{0.9221} & {32.36} & {0.9019} & {33.26} & \sotab{0.9378} & {39.35} & {0.9785} \\
    \arrayrulecolor{black!30}\midrule
{ATD-light}~\cite{zhang2024ATD}      & $\times$2 & 753K   & \sotaa{38.28} & \sotab{0.9616} & \sotab{34.11} & {0.9217} & \sotab{32.39} & \sotab{0.9023} & \sotab{33.27} & {0.9376} & \sotab{39.51} & \sotab{0.9789} \\
    \rowcolor{gray!15}
    \textbf{ATD-light+FKE}   & $\times$2 & 1163K   & \sotab{38.27} & \sotaa {0.9618} & \sotaa{34.26} & \sotaa{0.9229} & \sotaa{32.41} & \sotaa{0.9030} & \sotaa{33.55} & \sotaa{0.9405} & \sotaa{39.63} & \sotaa{0.9793} 
    \\
\arrayrulecolor{black!30}\midrule
    {ATD-light}~\cite{zhang2024ATD}   & $\times$3 & 760K   & \sotab{34.70} & \sotaa{0.9300} & \sotab{30.68} & \sotab{0.8485} & \sotab{29.32} & \sotab{0.8109} & \sotab{29.16} & \sotab{0.8710} & \sotab{34.60} & \sotab{0.9505} \\
    \rowcolor{gray!15}
    \textbf{ATD-light+FKE}   & $\times$3 & 1170K   & \sotaa{34.71} & \sotab{0.9298} & \sotaa{30.73} & \sotaa{0.8497} & \sotaa{29.35} & \sotaa{0.8120} & \sotaa{29.36} & \sotaa{0.8753} & \sotaa{34.69} & \sotaa{0.9513} \\
\arrayrulecolor{black!30}\midrule
    {ATD-light}~\cite{zhang2024ATD}      & $\times$4 & 769K   & \sotaa{32.62} & \sotaa{0.8997} & \sotab{28.87} & \sotab{0.7884} & \sotab{27.77} & \sotab{0.7439} & \sotab{26.97} & \sotab{0.8107} & \sotab{31.47} & \sotab{0.9198} \\
    \rowcolor{gray!15}
    \textbf{ATD-light+FKE}     & $\times$4 & 1179K   & \sotaa{32.62} & \sotab{0.8991} & \sotaa{28.93} & \sotaa{0.7902} & \sotaa{27.80} & \sotaa{0.7451} & \sotaa{27.12} & \sotaa{0.8163} & \sotaa{31.54} & \sotaa{0.9210} \\
    \arrayrulecolor{black}\bottomrule
  \end{tabular}
  \end{center}
\vspace{-1.em}
\end{table*}

\subsection{Comparisons with the state of the art}

\textcolor{violet}{\textbf{Group $\mathcal{A.}$}}
DeepRFTv2-B is trained on GoPro~\cite{Nah2017deep} and evaluated on GoPro~\cite{Nah2017deep} / HIDE~\cite{Shen2019human}. As shown in Table~\ref{tab:trained on GoPro}, DeepRFTv2 outperforms the other models in both PSNR and SSIM on the GoPro and HIDE test set. For example, DeepRFTv2-B achieves 34.52 / 32.22~dB on GoPro / HIDE test set, 0.46 / 0.48~dB higher than the kernel prior-assisted method UFPNet~\cite{fang2023UFPNet}. Fig.~\ref{fig:gopro_realblur} shows that our approach effectively minimizes a greater amount of blur degradation. 


\noindent\textcolor{violet}{\textbf{Group $\mathcal{B.}$}}
DeepRFTv2-B is trained and tested on RealBlur-R / J~\cite{Rim2020real} respectively. As seen in Table~\ref{tab:trained on RealBlur}, DeepRFTv2-B achieves 41.39~dB on RealBlur-R test set, 0.78~dB higher than UFPNet~\cite{fang2023UFPNet}. Specifically, when applied to RealBlur-J, DeepRFTv2-B achieves a superior result (34.46 dB) compared to UFPNet~\cite{fang2023UFPNet} (33.35 dB). As shown in Fig.~\ref{fig:gopro_realblur}. 
Additionally, we calculate PSNRs / SSIMs without aligning the deblurring result to the ground truth in Table~\ref{tab:trained on RealBlur} for further comparison. DeepRFTv2-B achieves 38.26~dB on RealBlur-R test set, 0.44~dB higher than EVSSM~\cite{kong2025EVSSM}. For RealBlur-J, DeepRFTv2-B achieves a better outcome (31.10~dB) than EVSSM~\cite{kong2025EVSSM} (30.68 dB). DeepRFTv2 achieves a significant improvement in the PSNR metric compared to other SOTA methods, regardless of the calculation method used.

\noindent\textcolor{violet}{\textbf{Group $\mathcal{C.}$}}
Moreover, DeepRFTv2-B achieves a competitive result with other methods for REDS-val-300~\cite{Nah2021ntire} shown in Table~\ref{tab:REDS-val-300}, \emph{e.g.} 0.22dB better than LoFormer~\cite{mao2024Loformer}. The quantitative experimental results indicate that our method also has a good ability to handle motion blur with JPEG artifacts.

\noindent\textcolor{violet}{\textbf{Group $\mathcal{D.}$}}
Our DeepRFTv2 also performs well for defocus deblur dataset DPDD~\cite{Abuolaim2020DPDD}. Table~\ref{tab:defocus} shows that DeepRFTv2-S  outperforms other SOTA methods in all evaluation metrics. Especially in terms of the perceptual evaluation metric (LPIPS), DeepRFTv2 achieves far superior results compared to other SOTA methods. In the combined scenes category, DeepRFTv2-S exhibits a 0.09 dB improvement over the leading method ALGNet~\cite{Gao2023ALGNet}. 

\noindent\textcolor{violet}{\textbf{Group $\mathcal{E.}$}} In order to further measure the effectiveness of FKE, we evaluate our FKE on the lightweight SR task which also can be explained by Eq.~\ref{eq:blur}. As indicated in Table~\ref{tab: results image sr2}, FKE can improve the performance of ATD~\cite{zhang2024ATD} from 33.27 / 29.16 / 26.97 dB to 33.55 / 29.36 / 27.12 dB on $\times2~/~3~/~4$ Urban100~\cite{Huang_2015_Urban100} benchmark. The results show that FKE also has a certain potential to handle other blur-related tasks. 

\begin{figure*}[t]
\begin{center}
    \includegraphics[width=0.95\linewidth]{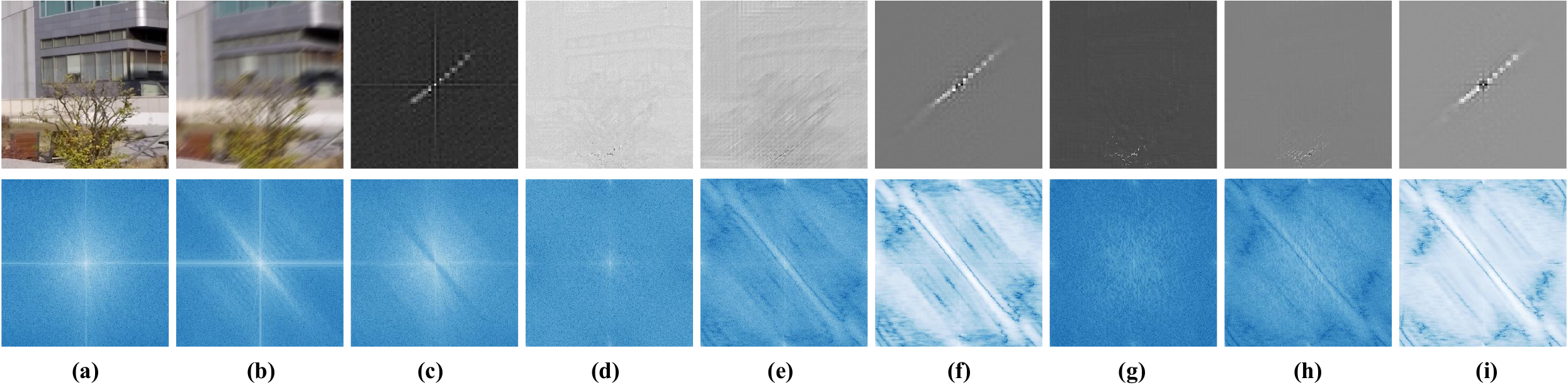}
\end{center}
\vspace{-1.2em}
\caption{Spatial and Fourier space visualizations of the features and the estimated kernels. (a) Sharp image; (b) Blur image; (c) ``Ground-truth'' blur kernel computed by $\mathcal{F}^{-1}(\mathcal{F}(\mathbf{B})/\mathcal{F}(\mathbf{S}))$. (d)$\sim$(f) and (g)$\sim$(i) Input feature, output feature and the corresponding estimated kernel by FKE, respectively.}
\label{fig:feature_kernel}
\vspace{-1.2em}
\end{figure*}
\section{Analysis and Discussion}
We have shown that DeepRFTv2 provides favorable results compared to other SOTA methods. In this section, we discuss the effectiveness of the main components. For the ablation studies in this section, all models are trained on the GoPro dataset using a batch size of 16 and a patch size of $256\times256$ for 200K iterations. 


\begin{table}[t] 
    \centering
\caption{ Ablation for different components. ``Res'', ``NAF'' and ``VS'' are Resblock, NAFBlock and VSBlock respectively. ``UNet'', ``MIMO'' and ``DMS'' are One-Scale UNet, MIMO-UNet and DMS-UNet architecture respectively. ``num'' and ``dim'' are the number and the channel number of UNet. {\color{red}\checkmark} means the quantity of basic block is doubled.}
\label{tab:ablation-1}
\vspace{-0.5em}
\renewcommand
\arraystretch{0.95}
\setlength{\tabcolsep}{1.9pt}
\resizebox{0.99\linewidth}{!}{
    \begin{tabular}{cc|ccc|ccccc|cc}
    \toprule[0.15em]
    \multicolumn{2}{c}{UNet} & \multicolumn{3}{c}{FKE} & \multicolumn{4}{c}{Architecture} & ~ \\
    \rowcolor{color4!50} NAF & VS & Res & NAF & VS & UNet & MIMO & DMS & num & dim & PSNR & FLOPs\\
     \midrule
     \checkmark & $\times$ & $\times$ & $\times$ & $\times$ & \checkmark & $\times$ & $\times$  & 1 & 56 & 33.00 & 34.93 \\
     \checkmark & $\times$ & $\times$ & $\times$ & $\times$ & $\times$ & \checkmark & $\times$  & 1 & 56 & 32.95 & 36.42 \\
     \checkmark & $\times$ & $\times$ & $\times$ & $\times$ & $\times$ & $\times$ & \checkmark  & 1 & 48 & 33.32 & 33.40 \\
     \arrayrulecolor{black!30}\midrule
     \checkmark & $\times$ & $\times$ & $\times$ & $\times$ & \checkmark & $\times$ & $\times$ & 1 & 48 & 32.74 & 25.81 \\
    \checkmark & $\times$ & \checkmark & $\times$ & $\times$ & \checkmark & $\times$ & $\times$ & 1 & 48 & 33.06 & 32.32 \\
    \checkmark & $\times$ & $\times$ & \checkmark & $\times$ & \checkmark & $\times$ & $\times$ & 1 & 48 & 33.10 & 28.82 \\
    \checkmark & $\times$ & $\times$ & $\times$ & \checkmark & \checkmark & $\times$ & $\times$ & 1 & 48 & 33.11 & 28.84 \\
    \checkmark & $\times$ & $\times$ & \color{red}\checkmark & $\times$ & \checkmark & $\times$ & $\times$ & 1 & 48 & 33.11 & 30.82 \\
    \checkmark & $\times$ & $\times$ & $\times$ & \color{red}\checkmark & \checkmark & $\times$ & $\times$ & 1 & 48 & 33.17 & 30.86 \\
    \arrayrulecolor{black!30}\midrule
    \checkmark & $\times$ & $\times$ & $\times$ & \checkmark & $\times$  & $\times$ & \checkmark  & 1 & 48 & 33.62 & 37.18 \\
    \checkmark & $\times$ & $\times$ & $\times$ &  $\times$ & $\times$  & $\times$ & \checkmark  & 2  & 48 & 33.66 & 67.19 \\
    \checkmark & \checkmark & $\times$ & $\times$ & \checkmark & $\times$  & $\times$ & \checkmark  & 1 & 48 & 33.84 & 59.67 \\
    \checkmark & $\times$ & $\times$ & $\times$ & \color{red}\checkmark & $\times$  & $\times$ & \color{red}\checkmark  & 1  & 48 & 33.90 &  70.26 \\
    \rowcolor{gray!15}\checkmark & $\times$ & $\times$ & $\times$ & \checkmark & $\times$  & $\times$ & \checkmark  & 2  & 48 & 34.10 &  75.34 \\
    \rowcolor{gray!15}\checkmark & \checkmark & $\times$ & $\times$ & \checkmark & $\times$  & $\times$ & \checkmark  & 2 & 48  & 34.34 &  120.3 \\
    \arrayrulecolor{black}\midrule[0.1em]
    \rowcolor{color4!50}- & - & - & - & - & - & - & - & - & rev &\multicolumn{2}{c}{GPU Memory}\\
    \arrayrulecolor{black}\midrule
    \checkmark & \checkmark & $\times$ & $\times$ & \checkmark & $\times$  & $\times$ & \checkmark  & 1 & $\times$ &\multicolumn{2}{r}{6,120 MB~~~~}\\
    \checkmark & \checkmark & $\times$ & $\times$ & \checkmark & $\times$  & $\times$ & \checkmark  & 2 & $\times$ &\multicolumn{2}{r}{12,027 MB~~~~}\\
    \checkmark & \checkmark & $\times$ & $\times$ & \checkmark & $\times$  & $\times$ & \checkmark  & 1 & \checkmark &\multicolumn{2}{r}{2,846 MB~~~~}\\
    \rowcolor{gray!15}\checkmark & \checkmark & $\times$ & $\times$ & \checkmark & $\times$  & $\times$ & \checkmark  & 2 & \checkmark &\multicolumn{2}{r}{3,439 MB~~~~}\\
    \arrayrulecolor{black}\bottomrule[0.15em]
    \end{tabular}}
    \vspace{-1.em}
\end{table}

\begin{table}[t]
    \centering
\small
\caption{ Ablation for Fourier Kernel Estimator. $\mathrm{B}$, $\hat{\mathrm{S}}$, $\mathrm{e_1}$ and $\mathrm{d_1}$ are blur image, restored image, encoder feature and decoder feature. ``Spatial'' and ``Fourier'' mean estimating kernels in the spatial domain (then convolved with the feature) and Fourier space, respectively. ``Act'' is applying ReLU, and ``Skip'' indicates the skip connection from $\mathrm{r}_i^j$ to $\mathrm{r}_{i+1}^{j-2}$. \textcircled{+} means replacing the multiplication with addition. }
\label{tab:ker1}
\vspace{-0.5em}
\renewcommand
\arraystretch{0.75}
\resizebox{0.95\linewidth}{!}{
    \begin{tabular}{cccc|cc|cc|c}
    \toprule[0.15em]
     $\mathrm{B}$ & $\hat{\mathrm{S}}$ & $\mathrm{e_1}$ & $\mathrm{d_1}$ & Spatial & Fourier & Act & Skip  & PSNR\\
     \midrule
     $\times$ & $\times$ & $\times$ & $\times$ & $\times$ & $\times$ & $\times$ & - & 32.74 \\
     $\times$ & $\times$ & $\times$ & \checkmark & \checkmark & $\times$ & \checkmark & - & 32.72\\
    $\times$ & $\times$ & $\times$ &  \checkmark & $\times$ & \checkmark & $\times$  & - &  32.74\\
     \arrayrulecolor{black!30}\midrule
    \checkmark & $\times$ & $\times$ & $\times$ & $\times$ & \checkmark & \checkmark & - & 32.79 \\
    $\times$ &  \checkmark & $\times$ & $\times$ & $\times$ & \checkmark & \checkmark & - & 32.84 \\
    $\times$ & $\times$ & \checkmark & $\times$ & $\times$ & \checkmark & \checkmark & - & 32.72 \\
    $\times$ & $\times$ & $\times$ &  \checkmark & $\times$ & \textcircled{+} & \checkmark & - & 32.90 \\
    $\times$ & $\times$ & $\times$ &  \checkmark & $\times$ & \checkmark & \checkmark & - & 33.06 \\
    \arrayrulecolor{black!30}\midrule
    \rowcolor{gray!15}$\times$ & $\times$ & $\times$ &  \checkmark & $\times$ & \checkmark & \checkmark & $\times$ & 34.04 \\
    \rowcolor{gray!15}$\times$ & $\times$ & $\times$ &  \checkmark & $\times$ & \checkmark & \checkmark & \checkmark & 34.10 \\
    \arrayrulecolor{black}\bottomrule[0.15em]
    \end{tabular}}
    \vspace{-1.em}
\end{table}

\subsection{Effectiveness of DMS-UNet}
To evaluate the efficacy of our DMS-UNet architecture, we constructed two baseline models: a One-Scale UNet and a MIMO-UNet, following the designs in Fig.~\ref{fig:DMSUNet}(b) and (c), respectively. As indicated in the first three rows of Table~\ref{tab:ablation-1}, 
our DMS-UNet achieves superior performance (33.32 dB PSNR) compared to both the One-Scale UNet (33.00 dB) and the Multi-Scale UNet (32.95 dB), while also requiring lower computational cost. 

Compared with One-Scale UNet with FKE (33.11 dB 7th row), DMS-UNet with FKE achieves a better performance (33.62 dB 10th row). Additionally, the architecture of multiple reversible sub-unets (34.10 dB 14th row) performs much better than a single large UNet (33.90 dB 13th row). Meanwhile, NAFEVSBlock can increase the performance from 34.10 dB (14th row) to 34.34 dB (15th row).

Moreover, GPU memory is calculated by training a single patch size 256×256 image. As shown in the last five rows of Table~\ref{tab:ablation-1}, our reversible architecture (Revversible-UNet and Reversible-ResNet) can effectively save the required training memory. The memory consumption of a non-reversible architecture increases at a much faster rate than that of a reversible architecture as the number of DMS-UNets rises.

\subsection{Effectiveness of Fourier Kernel Estimator}
The core of our proposed method is to allow the neural network to predict the kernels without requiring additional supervision in Fourier space. Table~\ref{tab:ker1} shows that estimating kernels in the spatial domain (32.72 dB, $2$nd row, Fig~\ref{fig:ablation_kernel} (a)) does not work, which behaves exactly the same as if no Kernel Estimator was added (32.74 dB, $1$st row), while estimating kernels in Fourier space (33.06 dB, $8$th row) yields a 0.32 dB improvement.
Meanwhile, without the activation process in Fourier space, kernel estimation process does not work (32.74 dB, $3$rd row, Fig.~\ref{fig:ablation_kernel} (b)). 

FKE is located in the tail part of each sub-unet, taking the decoder feature $\mathrm{d_1}$ as input. We further put FKE in other parts of the sub-unet and change the input features from $\mathrm{d_1}$ to other alternatives (see Fig.~\ref{fig:rev-unet} (b)), which include: the blur image $\mathrm{B}$, the latent sharp image $\hat{\mathrm{S}}$ (generated by $\mathrm{d_1}$) and the encoder feature $\mathrm{e_1}$. The results indicate that FKE works best with the decoder feature $\mathrm{d_1}$, which has been fully extracted through the spatial network (mid-region in Table~\ref{tab:ker1}). This shows that FKE requires features from a later stage of the network to learn an effective representation of the blur process (see more insight analysis in Sec.~\ref{sec:KlL}). If we cancel the skip-connection in the FKE of DeepRFTv2 (34.10 dB, last row), the performance will drop to 34.04 dB. Compared to the model without FKE (33.66 dB, 67.19 G in Table~\ref{tab:ablation-1}), FKE brings a significant performance growth at a low computational cost (34.10 dB, 75.34 G). If we change multiplication in Fourier space with addition (like DeepRFT~\cite{XintianMao2023DeepRFT}), it can still learn some kernel pattern (shown in Fig.~\ref{fig:ablation_kernel} (c)) but it will lead to a performance drop. This is because the addition mode does not let the model learn the convolution process.


Furthermore, we select three blocks as candidate basic modules: ResBlock~\cite{He2016ResNet}, NAFBlock~\cite{Chen2022simple} and VSBlock~\cite{kong2025EVSSM}. Results are shown in Table~\ref{tab:ablation-1} ($4\sim9$th rows). Compared with NAFBlock~\cite{Chen2022simple} and VSBlock~\cite{kong2025EVSSM}, ResBlock obtains lower performance with higher computational cost. Although NAFBlock (33.10 dB) performs as well as VSBlock (33.11 dB), as the network deepens, its performance improvement (33.11 dB) is not as significant as VSBlock's (33.17 dB). 



\subsection{Kernel-level Learning}
\label{sec:KlL}
Generally, neural networks do not have the ability to learn kernels in the spatial domain. As shown in Fig.~\ref{fig:feature_kernel} (d) and (g), they are often good at extracting texture features rather than the blur features. As can be seen in Fig.~\ref{fig:feature_kernel} (e) and (h), FKE enables the model to explicitly learn the essence of blur by directly convolving the features with the predicted kernels. 

\begin{figure}[t]
\centering
 \includegraphics[width=1.0\linewidth]{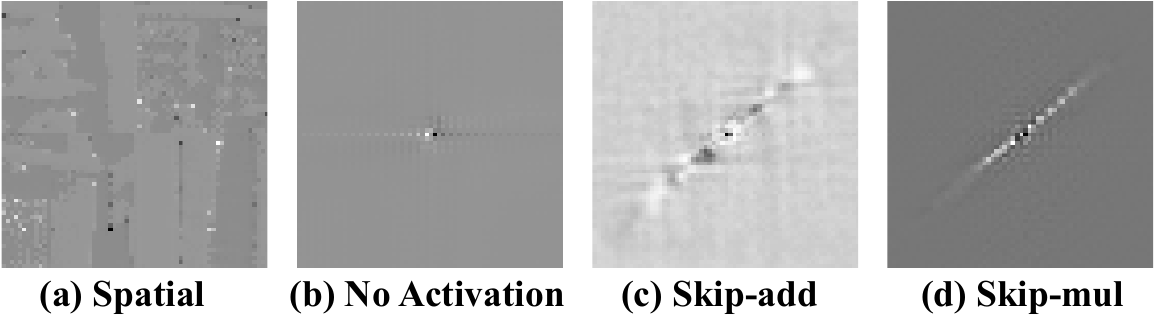}    
\vspace{-1.5em}     
\caption{ Estimated kernels in various modes. (a) Estimate kernels and convolve them with features in the spatial domain. (b) Remove ReLU in FKE. (c) Skip-add kernels with features in FKE. (d) The proposed FKE.}
\label{fig:ablation_kernel}
\vspace{-1.2em}
\end{figure}

\begin{figure}[t]
\centering
 \includegraphics[width=1.0\linewidth]{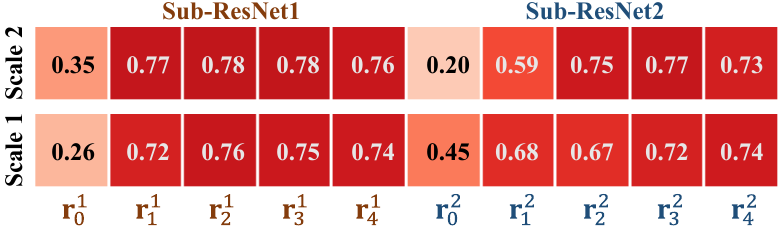}    
\vspace{-1.5em}     
\caption{CKA similarities~\cite{kornblith2019CKA} of the FKEs' features and the ``ground-truth'' blur kernel (with noise) in Fourier space.}
\label{fig:cka_reblurker}
\vspace{-1.4em}
\end{figure}

As can be seen in Fig.~\ref{fig:feature_kernel} (f) and (i), the kernels predicted by our FKE are strongly correlated with the real blur kernels (Fig.~\ref{fig:feature_kernel} (c)). Other modes (shown in Fig.~\ref{fig:ablation_kernel}) do not enable the network to effectively perform kernel learning. Furthermore, we calculate the CKA similarities~\cite{kornblith2019CKA} of the features and the blur kernel in Fig.~\ref{fig:cka_reblurker}. The results show that the features are close to the real blur kernel, while there are still certain differences. This is due to the fact that the supervision specifically for the kernel estimator to reblur the sharp image is not incorporated, thus enlarging the dynamic space for kernel estimation. 

Based on the above analysis, without any supervision, our FKE can estimate a practically meaningful kernel by instantiating the physical blur process. One primary reason for causing this phenomenon stems from the activation in Fourier space which has been analyzed in our DeepRFT~\cite{XintianMao2023DeepRFT}. As can be observed in Fig.~\ref{fig:fft_relu}, after FFT-ReLU-IFFT, half-image subtraction and cyclic shift, the results strikingly resemble the blur direction. As a result, the feature $\mathrm{r}_1$ closely approximates the blur kernel, whereas the shallow feature $\mathrm{r}_0$ does not (shown in Fig.~\ref{fig:cka_reblurker}). 

The experiment and visualization demonstrate that the activation in Fourier space is one of the keys to kernel estimation. Based on the activation in Fourier space, the FKE estimates practical significant kernels for convolution. It also enables the network to explicitly learn the essence of blur, which is the convolution of a sharp image with a blur kernel.
As shown in Fig.~\ref{fig:feature_kernel}, compared with global kernels at the image-level which is vulnerable to from noise and spatially-variant blur problem, the feature-level one demonstrates better flexibility. FKE supplements the blur information of the loss in the pixel-level learning process of the network by introducing kernel-level learning. 


\section{Conclusion}
\label{sec:conclusion}
{
In this paper, we present DeepRFTv2, a novel end-to-end image deblurring model. The core contribution is the Fourier Kernel Estimation (FKE), which is integrated with the Decoupled Multi-Scale UNet (DMS-UNet) to achieve robust performance. The FKE leverages the properties of Fourier space to reformulate the challenging kernel estimation problem into a more tractable multiplicative matrix prediction task. Unlike previous methods that rely on separate supervision, our FKE is embedded within the end-to-end framework. This integration enhances the adaptability of the estimated kernels, allowing them to function as real global convolution operators, which enables the model to learn the blur process in kernel-level. Furthermore, the proposed DMS-UNet addresses the issue of information aliasing in conventional MIMO-UNet by facilitating bidirectional multi-scale information flow across its hierarchical sub-unets, thus improving deblurring performance. Extensive experiments on benchmark datasets establish that our method achieves state-of-the-art performance in motion deblurring, demonstrating considerable potential for other kernel-related tasks.
}

{
\small
\bibliographystyle{IEEEtran}
\bibliography{./egbib.bib}
}

\begin{IEEEbiography}[{\includegraphics[width=1in,height=1.25in,clip,keepaspectratio]{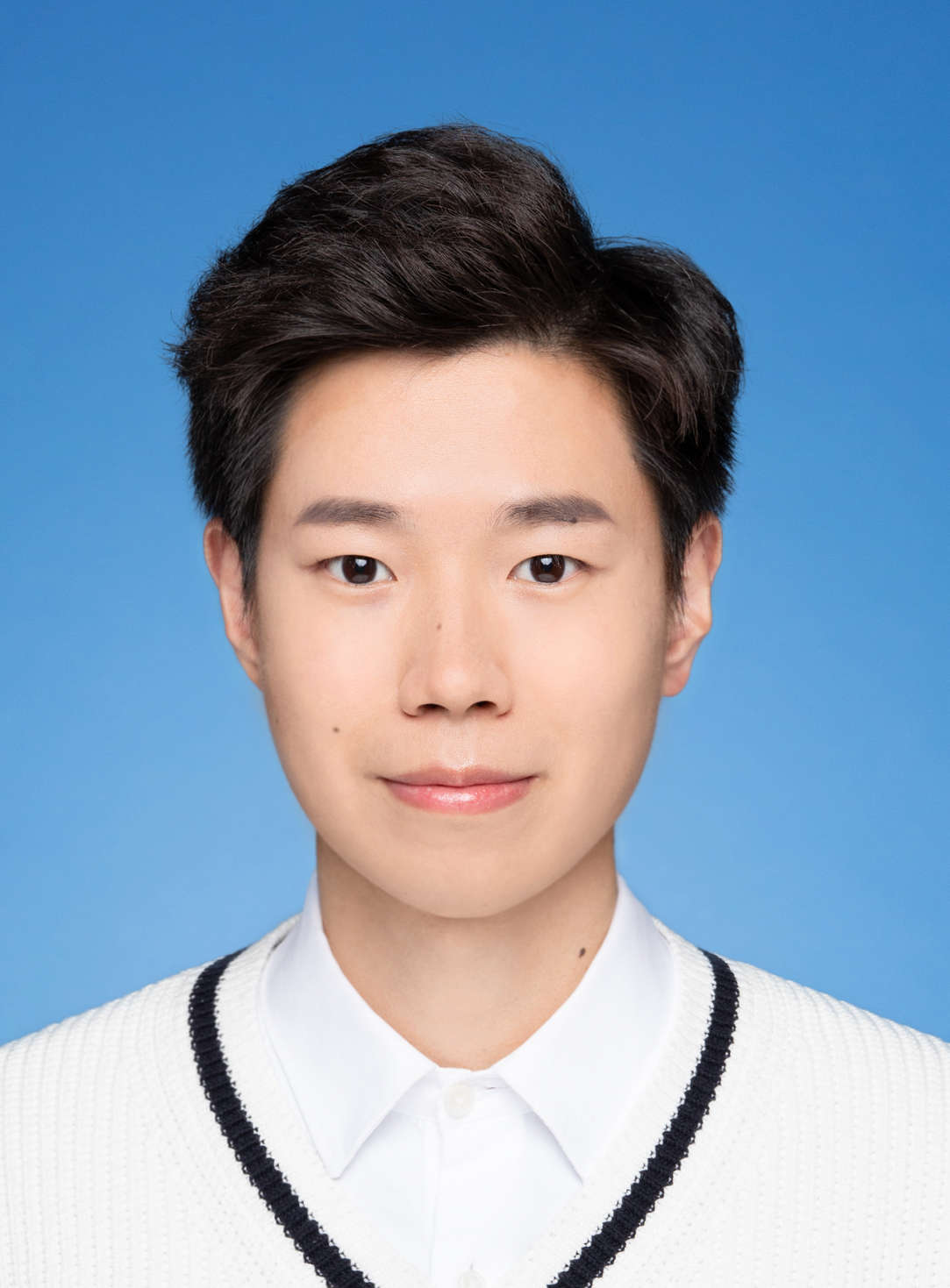}}]{Xintian Mao}
received the B.S. degree in Electronic Information Science and Technology from East China Normal University, Shanghai, China, in 2019, and the M.S. degree in Communication and Information System from East China Normal University, Shanghai, China, in 2022. He is currently Eng.D. at Shanghai Key Laboratory of Multidimensional Information Processing, the School of Communication and Electronic Engineering, East China Normal University. His research interests include image deblurring, medical image restoration,  and remote sensing.
\end{IEEEbiography}

\begin{IEEEbiography}[{\includegraphics[width=1in,height=1.25in,clip,keepaspectratio]{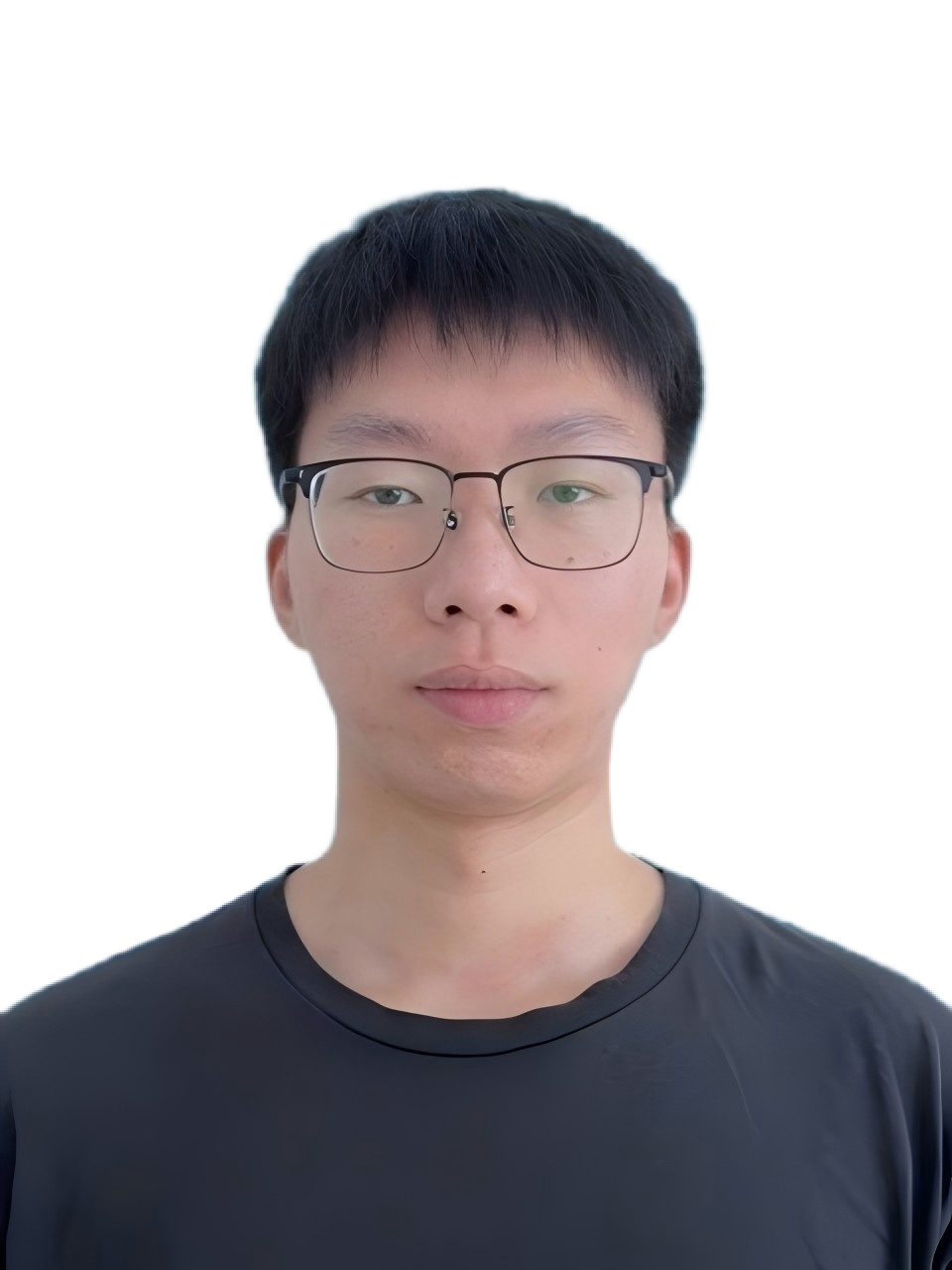}}]{Haofei Song}
received the B.S. degree in Automation from East China University of Science and Technology, Shanghai, China, in 2022, and the M.S. degree in Information and Communication Engineering from East Chian Normal University, Shanghai, China, in 2025. He is currently Ph.D. at Shanghai Key Laboratory of Multidimensional Information Processing, the School of Communication and Electronic Engineering, East China Normal University. His research interests include image restoration, open-vocabulary learning and medical image processing.
\end{IEEEbiography}

\begin{IEEEbiography}[{\includegraphics[width=1in,height=1.25in,clip,keepaspectratio]{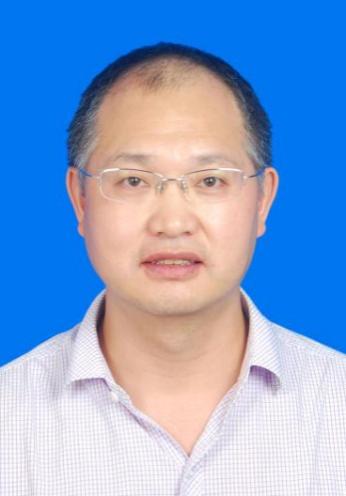}}]{Yin-Nian Liu}
(Member, IEEE) received the B.S. degree in physics from Northwest Normal University, Lanzhou, China, in 1994, the M.S. degree in physics from the Xi’an Institute of Optics and Precision Mechanics, Chinese Academy of Sciences, Shanghai, China, in 1997, and the Ph.D. degree in remote sensing from the Shanghai Institute of Technical Physics, Chinese Academy of Sciences, in 2005.

He is currently a Senior Researcher with the Shanghai Institute of Technical Physics. He is a Principle Investigator of the Advanced Hyperspectral Imager (AHSI) on the GaoFen-5 (GF-5), GF5–02, ZiYuan-1-02D (ZY1-02D), and ZY1-02E satellites. His research interests include infrared and hyperspectral imaging technology, high-precision calibration technology, and remote sensing image applications. His research interests also include hyperspectral anomaly detection and image processing.,Dr. Liu received the China National Sci-Tech Prize in 2012, the Shanghai Sci-Tech Prize in 2009, and the Award of Shanghai Leading Talents in 2014.
\end{IEEEbiography}

\begin{IEEEbiography}[{\includegraphics[width=1in,height=1.25in,clip,keepaspectratio]{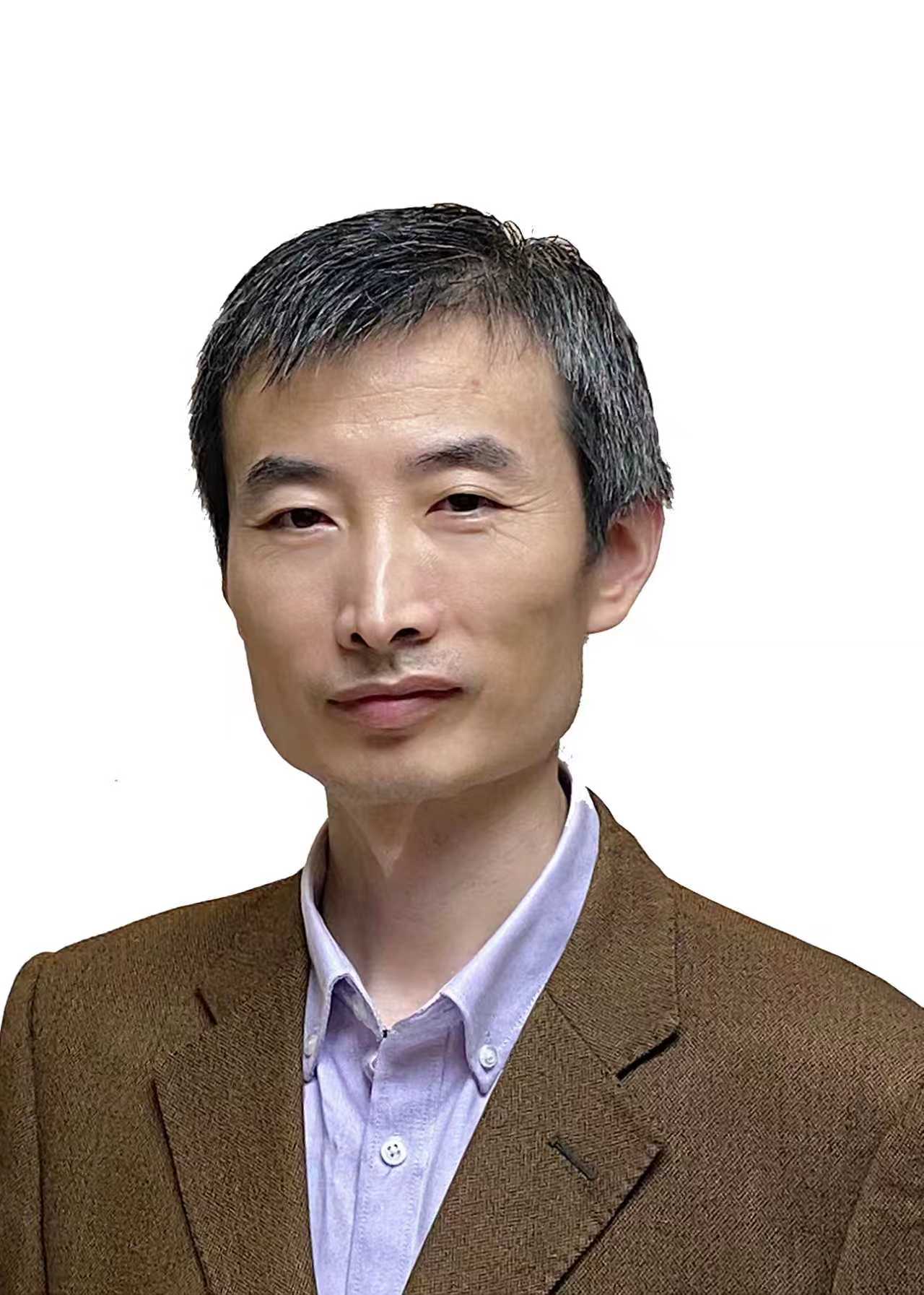}}]{Qingli Li}
 received the B.S. and M.S. degrees in computer science and engineering from Shandong University, Jinan, China, in 2000 and 2003, respectively, and the Ph.D. degree in pattern recognition and intelligent system from Shanghai Jiao Tong University, Shanghai, China, in 2006. From 2012 to 2013, he was a visiting scholar at Medical Center, Columbia University, New York, USA. He is currently with the Shanghai Key Laboratory of Multidimensional Information Processing, East China Normal University, Shanghai, China. His research interests include molecular imaging, biomedical optics, and pattern recognition. He is a senior member of IEEE and member of SPIE. He is an editorial board member of International Journal of Computer Assisted Radiology and Surgery.
\end{IEEEbiography}

\begin{IEEEbiography}[{\includegraphics[width=1\linewidth]{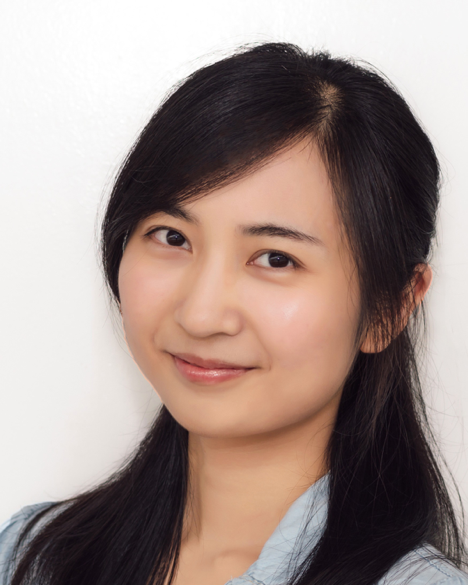}}]{Yan Wang} received her B.E. degree from Huazhong University of Science and Technology, China, and completed her Ph.D. degree at Nanyang Technological University, Singapore. She was a research scholar at the Ohio State University. She then held a postdoctoral position with the Center for Imaging Science, the Johns Hopkins University (2017-2020). She is currently a Professor at Shanghai Key Laboratory
of Multidimensional Information Processing, the School of Communication and Electronic Engineering at East China Normal University. Her research interests lie in the fields of medical image analysis and hyperspectral imaging. She served as an area chair for CVPR, MICCAI, and an associate editor for multiple journals. 
\end{IEEEbiography}


\end{document}